\documentclass[12pt, letterpaper]{article}

\usepackage[margin=1in]{geometry}
\usepackage[utf8]{inputenc}

\usepackage{amssymb}
\usepackage{amsmath}
\usepackage{lineno}
\usepackage{multirow}
\usepackage{graphicx}
\usepackage{subcaption}
\usepackage{arydshln}

\usepackage{xcolor}

\usepackage[authoryear]{natbib}
\usepackage{hyperref}

\usepackage{tikz}
\usepackage{caption}
\usetikzlibrary{shapes.geometric, arrows.meta, positioning}

\tikzstyle{block} = [rectangle, draw, text centered, minimum height=2em, text width=4.5cm]
\tikzstyle{phase} = [draw=none, font=\bfseries, text centered]
\tikzstyle{arrow} = [thick,->,>=stealth]
\tikzstyle{dashedarrow} = [thick,dashed,->,>=stealth]

\title{Machine Learning Workflows in Climate Modeling: Design Patterns and Insights from Case Studies\footnote{This research is supported by funding from the National Science Foundation of the U.S.A. through the Learning the Earth with Artificial Intelligence and Physics (LEAP) Science and Technology Center (STC)(Award No.2019625).}}

\author{
	Tian Zheng$^{1,2,\dagger}$, Subashree Venkatasubramanian$^{2}$, Shuolin Li$^{2,3}$, \\Amy Braverman$^{4}$, Xinyi Ke$^{1,2}$, Zhewen Hou$^{1}$, \\Peter Jin$^{5}$, Samarth Sanjay Agrawal$^{2}$}

\date{\small $^{1}$Department of Statistics, Columbia University, New York, New York; \\
	$^{2}$ NSF STC Learning the Earth with AI and Physics (LEAP), New York, New York; \\
	$^{3}$ Data Science Institute, Columbia University, New York, New York; \\
	$^{4}$ Jet Propulsion Laboratory, California Institute of Technology, Pasadena, California;\\
	$^{5}$ Department of Applied Physics and Applied Mathematics, Columbia University, New York, New York;  \\
	$\dagger$ Correspondence: tian.zheng@columbia.edu.}

\begin{document}

\maketitle

\begin{abstract}
	Machine learning has been increasingly applied in climate modeling on system emulation acceleration, data-driven parameter inference, forecasting, and knowledge discovery, addressing challenges such as physical consistency, multi-scale coupling, data sparsity, robust generalization, and integration with scientific workflows. This paper analyzes a series of case studies from applied machine learning research in climate modeling, with a focus on design choices and workflow structure. Rather than reviewing technical details, we aim to synthesize workflow design patterns across diverse projects in ML-enabled climate modeling: from surrogate modeling, ML parameterization, probabilistic programming, to simulation-based inference, and physics-informed transfer learning. We unpack how these workflows are grounded in physical knowledge, informed by simulation data, and designed to integrate observations.  We aim to offer a framework for ensuring rigor in scientific machine learning through more transparent model development, critical evaluation, informed adaptation, and reproducibility, and to contribute to lowering the barrier for interdisciplinary collaboration at the interface of data science and climate modeling.
\end{abstract}


\section{Introduction}
Machine learning (ML) is playing an increasingly important role in climate science \citep{balaji_climbing_2021}, from accelerating simulations to enabling new forms of prediction and system emulation \citep{rasp2018deep,chen_iterative_2023}. Yet, integrating machine learning into climate modeling presents distinct challenges, particularly around preserving physical relevance, handling sparse or heterogeneous data, and ensuring workflow transparency and reliability. Machine learning is not used as a generic plug-in to existing models. Rather, a successful machine learning application in climate modeling hinges on thoughtful workflow design. This includes not just model architecture, but also decisions about problem setup, data preparation, training procedures, evaluation diagnostics, and iteration cycles. 

Exciting adoption of machine learning into climate modeling is happening broadly to address important scientific problems \citep{eyring2024pushing}. As a result, a rich and growing literature has emerged, with the specific applications of machine learning and statistics scattered and often buried within domain-specific discussions of results, and technological terms vary across studies. As a result, it could be challenging for readers coming from a statistical and machine learning research perspective to compare papers with similar end goals in terms of what design choices were made, to reproduce results, to evaluate the contributions of particular machine learning components, or to scrutinize assumptions in ways that might guide further research.

Tukey (1962) \citep{tukey1962future} shared his belief that statisticians should analyze \textit{data analysis practices} of others to identify evolving methodological trends. Donoho (2017) \citep{Donoho_2017} argued in 2017 that cross-study and cross-workflow analyses, enabled by \textit{transparent and reproducible data science research practices}, will become commonplace in science in the future. In this paper, we unpack and synthesize the human design decisions that shape ML-enabled climate modeling workflows using a common machine learning framework, focusing on what data are used and how they are produced, how models are constructed, trained, and validated, and what compromises are made and why. Furthermore, modular workflows support rigorous methodology evaluation. Transparent documentation of these choices is critical for ensuring reproducibility, enabling critical evaluation, driving further methodological research, and supporting interdisciplinary collaboration. Many have advocated for this need for transparency in ML-enabled data science. A 2014 editorial in Science \citep{mcnutt2014reproducibility} highlighted the broader scientific importance of making methodological choices transparent. Such concerns have driven efforts in natural and social sciences to establish reproducible and transparent data science practices. For example, Mitchell et al.\ (2019) \citep{mitchell2019model} proposed \textit{Model Cards} to document key modeling decisions; Yu and Kumbier (2020) \citep{Yu_Kumbier_2020} introduced Veridical Data Science principles to guide the evaluation of human choices. More recently, Kapoor and Narayanan (2023) \citep{kapoor2023leakage} highlighted the problem of data leakage as a key driver of irreproducibility in ML-based science and offered a \textit{model info sheet} template to detect and address such leakages.

This paper offers the perspective of machine learning researchers working at the interface of statistics, machine learning, and climate modeling, showing how machine learning design choices are driven by scientific problems, existing scientific workflows, available data, and computational constraints. It is intended primarily for data scientists and machine learning researchers who wish to engage in climate applications, and secondarily for climate scientists seeking to understand, evaluate, and adapt machine learning workflows. By clarifying terminology, exposing modular design patterns, and demonstrating a framework for transparent discussion of choices with rich case studies, we aim to lower barriers to interdisciplinary collaboration, improve reproducibility and transferability, and support the development of scientifically robust machine learning workflows. 

Readers seeking broader perspectives on the emerging role of artificial intelligence and machine learning in shaping future climate modeling are referred to excellent position papers in recent literature \citep{rolnick2022tackling,eyring2024pushing,gettelman2022future}.

We begin in Section 2 with a conceptual overview of how physics, data, and machine learning interact in climate modeling. Section 3 uses a common machine learning workflow framework to thread our discussion of design choices from case studies. Full case studies are summarized using the framework in a supplement. We conclude in Section 4 with a discussion of open challenges and opportunities for further research on reproducibility and online evaluation, scalability, and integration with scientific inquiries.

\section{Physics+Data+ML: ML-Enabled Climate Modeling}
\label{sec:triangle}

\paragraph{Background on Climate Modeling. }
Climate models, particularly Earth System Models (ESMs), simulate the evolution of the Earth's atmosphere, ocean, land, and cryosphere by numerically solving a set of fundamental \textit{physical equations} \cite{gettelman2016demystifying}. These models discretize the Earth into a three-dimensional \textit{grid}, resolving the dynamics of fluid motion and thermodynamics through time. Key variables like temperature, wind, pressure, and humidity are calculated at each grid point by solving the Navier–Stokes equations, radiative transfer, and energy balance equations. For a comprehensive and accessible introduction to this topic, see \cite{gettelman2016demystifying}.

However, certain \textit{climate processes} cannot be explicitly modeled, either because they operate below the resolution of the \textit{model grid} or lack a well-defined physical formulation. Phenomena such as cloud formation, convection, and turbulent mixing occur at scales finer than typical grid resolutions and are therefore represented through \textit{parameterizations}, which are empirical or statistical representations that approximate the aggregate influence of \textit{unresolved} small-scale processes on the larger-scale state variables resolved by the model\citep{gentine2018could}. \textit{Model tuning} plays a crucial role in aligning model output with observed behavior by adjusting parameters to satisfy global constraints \cite{hourdin2017art}. Traditional \textit{data assimilation} techniques further refine short-term forecasts by nudging model states toward observations, but these approaches struggle to manage structural biases, observational uncertainty, and long-range variability \cite{gettelman2022future}.

\textit{Reanalysis datasets} (such as ERA5 \cite{hersbach_era5_2020}) are one response to these limitations. They combine historical observations with physics-based models through data assimilation to produce gridded, physically consistent estimates of atmospheric and oceanic variables over time. These datasets are critical for evaluating models, initializing simulations, and training machine learning systems that require spatiotemporally continuous inputs.

\begin{figure}[h]
	\centering
	\includegraphics[width=0.8\linewidth]{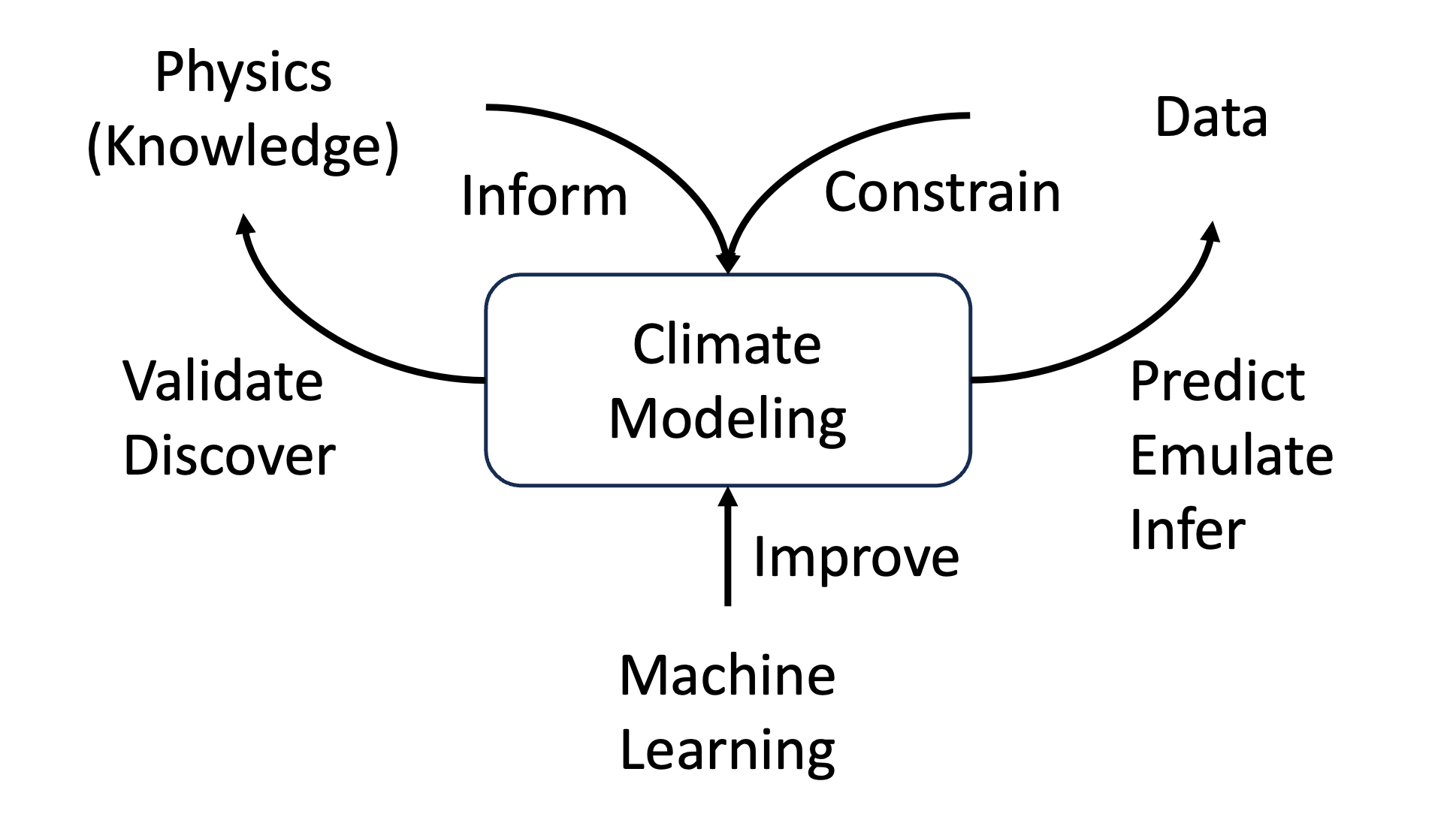}
	\caption{Interacting roles of physics, data, and machine learning in climate modeling. Climate models, especially Earth System Models (ESMs), interact with two foundational sources of information: physics and data (simulated or observed). Machine learning is increasingly used to improve the models' reliability and utility. Physics informs model structure, while models in turn help discover new scientific insights. Data constrain model behavior, and models help emulate climate data, interpret observations, and infer unmeasured physical states. Machine learning contributes by improving model components and enabling new approaches to represent the climate system’s variability and dynamics.}
	\label{fig:ml-climate-triangle}
\end{figure}

\paragraph{ML-Enabled Climate Modeling.}
As the societal need for more reliable climate projections increases, climate modeling faces a growing set of bottlenecks: increasing model complexity, rising computational costs, persistent structural biases, and limits in observational data \cite{balaji_are_2022}. Conventional Earth system models are also limited in their capabilities in leveraging the growing availability of observational datasets as new opportunities for model evaluation, training, and data-driven discovery. At the same time, emerging computational architectures and machine learning (ML) methods offer new tools for model acceleration, uncertainty quantification, and data-driven research \cite{baker_workshop_2019}. Machine learning introduces powerful function approximation, pattern recognition, and inference capabilities. When integrated with physics-based ESMs and observational data, machine learning has the potential to transform climate modeling workflows. Rather than replacing physics-based models, machine learning can be used to improve them: emulating expensive components, extracting structure from data, and enabling faster and more adaptive workflows.

As machine learning becomes increasingly embedded in climate modeling, it is clear that predictive performance and scientific insight depend not on machine learning alone, but on the thoughtful integration of three foundational elements: physical theory, data, and machine learning. As Gettleman et al.\ (2022) \cite{gettelman2022future} note, advancing predictive skill requires "a more holistic, integrated, and coordinated approach to models, observations, and their uncertainties." This triadic relationship—physics, data, and ML—is central to emerging climate modeling workflows.

This section uses the conceptual diagram in Figure~\ref{fig:ml-climate-triangle} to illustrate these interconnections. Each edge represents a relationship: models are guided by physics, constrained and validated by data, and increasingly augmented by machine learning for flexibility and scalability.

The interplay between prediction, understanding, and data-driven learning is not new \cite{balaji_climbing_2021}. The current mix of enthusiasm and concerns in the climate modeling community towards machine learning resembles that which arose in an earlier era, where meteorologists debated whether forecasting should rely on theory-based simulation or empirical pattern recognition. This time, however, machine learning may offer opportunities to combine them \cite{gettelman2022future, balaji_climbing_2021,balaji_are_2022}.

Following the diagram in Figure~\ref{fig:ml-climate-triangle}, we structure ML-enabled climate modeling into three integration styles, each defined by its starting point and its primary emphasis in the physics–data–ML triad. Progress is being made in all three directions, and they are not mutually exclusive as shown in Figure~\ref{fig:ml-climate-triangle}.

\paragraph{Direction I: Physics-First and ML-Augmented Simulation-Based Modeling.}
These workflows retain physics-based models at the core and use machine learning to accelerate components, discover new structure, or improve parameterizations. They seek to preserve theoretical grounding while extending computational feasibility and flexibility. While machine learning surrogates of computation-intensive components of Earth system models can lead to improved computational efficiency, key challenges in this direction include stability of hybrid models, trustworthiness of learned components, and seamless integration with legacy simulation codes. Case studies discussed in Section 3 (with full summaries in Supplement Section~\ref{sec:physicsfirst}) include surrogate modeling using neural operators \cite{kovachki2023neural} to improve system emulation (Case Study 1a: Neural Operators), equation discovery \cite{rackauckas2020universal,brunton2016discovering} to refine and simplify physical laws (Case Study 1c: Equation Discovery), and a systematic benchmarking of data preparation and model choices' impacts on ML parameterizations (Case Study 1d: Benchmarking ML Parameterization Workflows). We also describe a recent effort to integrate ML-trained surrogate models into climate models using Docker environments systematically \cite{bocquet2023surrogate} and Ftorch \cite{atkinson2025ftorch}, thereby bridging Python machine learning workflows with Fortran-based simulations (Case Study 1b: ClimSim). These approaches aim to reduce computational cost while improving physical realism and enabling large ensemble simulations. 

\paragraph{Direction II: Data-First (Observation-Integrated) ML Modeling.}
These workflows begin with observations, leveraging machine learning to extract structure, perform inference, and integrate observational data with process understanding. They often focus on calibrating and validating models, especially where theoretical understanding is limited or indirect. In Section 3 and Supplement Section~\ref{sec:datafirst}, we present one case study focuses on simulation-based inference \cite{PNASCranmerBrehmerAndLouppe2020} for remote sensing: using forward simulations with machine learning modules to make sense of satellite observations and carry out statistical inference on latent physical quantities (Case Study 2a: Uncertainty Quantification for Remote Sennsing). A second case study \cite{li2025probabilistic} in Supplement Section~\ref{sec:datafirst} uses probabilistic programming to infer model parameters from long-term stationary distributions derived from observations (Case Study 2b), highlighting how machine learning can be used for scientifically meaningful inverse problems. These approaches complement or extend traditional data assimilation by incorporating more flexible and expressive machine learning methods. Key challenges in this direction include data availability, temporal non-stationarity, and measurement uncertainty, all of which impact generalization.

\paragraph{Direction III: ML-First Climate Modeling.}
These workflows aim to build ML-based models that emulate or predict climate system behavior with minimal reliance on explicit physical equations. They prioritize predictive accuracy over theoretical interpretability, often leveraging large-scale datasets. One case study in Supplement Section~\ref{sec:mlfirst} benchmarks a range of machine learning architectures, from lightweight models such as ResNet \cite{ResNetRaspThuerey2021} to foundation models such as FourCastNet \cite{pathak2022fourcastnet}, for subseasonal to seasonal forecasting \cite{nathaniel2024chaosbench}, focusing on skill over interpretability (Case Study 3a: ChaosBench Benchmarking for Subseasonal Forecasting). Supplement Section~\ref{sec:mlfirst} unpacks another case study \cite{kim_spatiotemporal_2024} that develops a model to extrapolate sparse air–sea flux observations across the global ocean (Case Study 3b: Transfer Learning for pCO$_2$ Upscaling). The model is first trained on simulated data and then fine-tuned with real observations using a transfer learning \cite{pan2010domain} framework. These models are typically designed for generalization, speed, and data fusion, not mechanistic understanding. Key challenges in this direction include data availability for modeling training, long-term generalization, and the absence of built-in physical constraints. This direction also raises foundational questions about the meaning and limits of climate simulation using learned systems, which is outside the scope of this paper.

\section{Structuring Machine Learning Workflows for Climate Modeling with Science-Guided Design Choices}

In addition to the three directions in Figure~\ref{fig:ml-climate-triangle}, a fourth "direction" is human-in-the-loop: the design of workflows guided by scientific reasoning, modeling practices, and domain expertise. Reliable machine learning workflows in this setting must support interpretability, transparency, and relevance to physical understanding \cite{Donoho_2017,Yu_Kumbier_2020}. To achieve these goals, workflows should be designed and reviewed in terms of modular components and supported by diagnostic tools that facilitate iteration, validation, and integration with scientific implementation. Throughout this section, we use case studies to illustrate how design choices arise in practice and identify recurring patterns.

\begin{figure}[h]
	\centering
	\includegraphics[width=\linewidth]{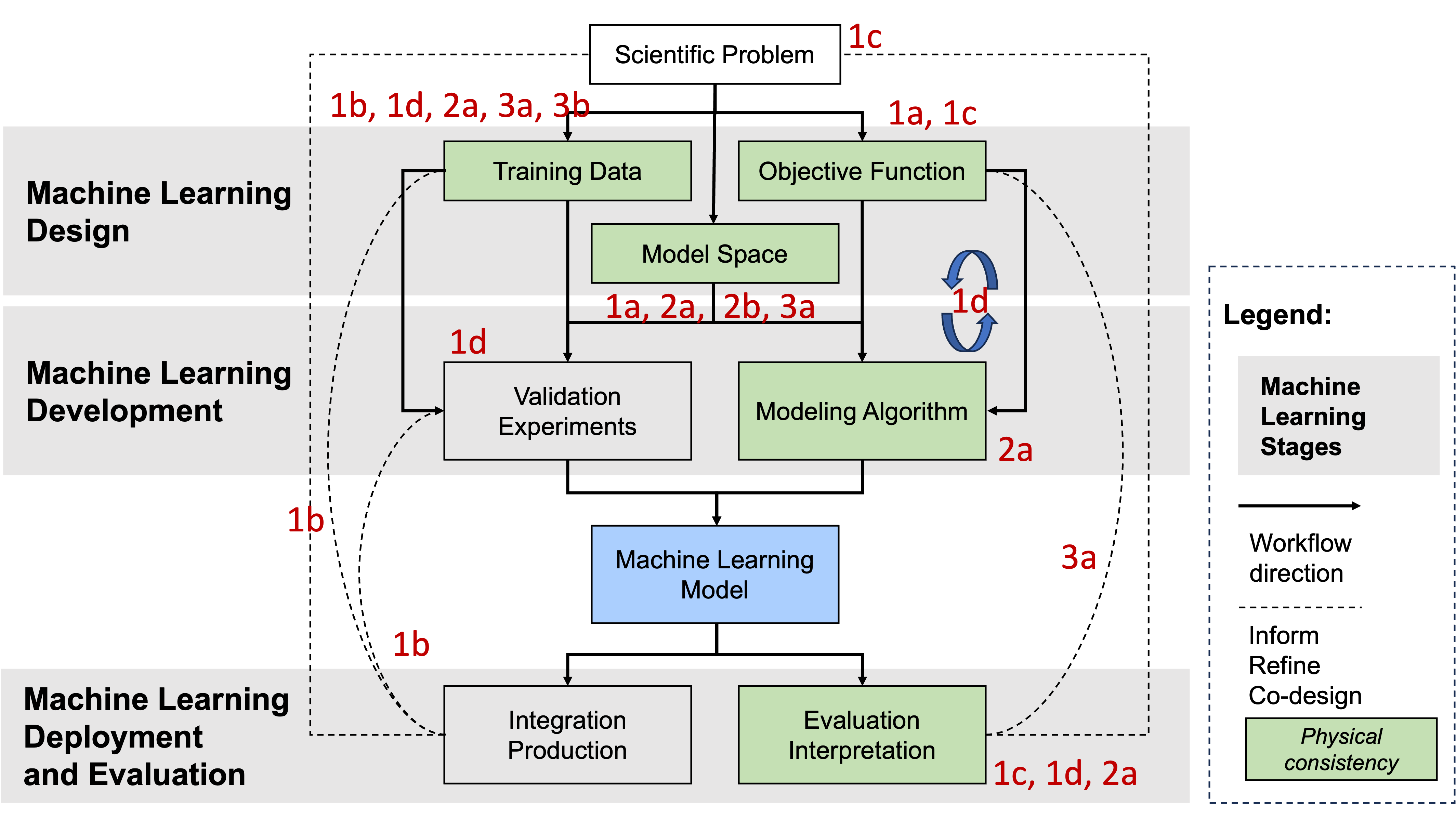}
	\caption{Machine learning workflow for climate modeling. Applying machine learning to Earth system modeling and climate data analysis spans three phases: design, development, and deployment. At each stage, domain knowledge and physical consistency guide decisions about model structure, objective functions, and evaluation. The process often iterates between design and development. Model integration, evaluation, and interpretation then lead to operational and scientific insights, supporting adaptive problem refinement and co-design in the workflows. Case studies that provide in-depth discussions of design choices are labeled in red (case study ID). This diagram is not meant to be a novel contribution, but rather to offer a familiar machine learning framework used to organize and contextualize our discussion.}
	\label{fig:ML-workflow}
\end{figure}

In this paper, we focus primarily on supervised or predictive machine learning workflows, as opposed to exploratory or unsupervised approaches (with exception of Case Study 2a). To structure our discussion, we adopt a common machine learning workflow framework (Figure~\ref{fig:ML-workflow}) that reflects widely recognized stages in scientific machine learning. This framework is not intended as a novel contribution; rather, it serves as a familiar structure to thread our analysis of design choices in ML-enabled climate modeling workflows. We further use this framework to structure full case study summaries in the Supplement, similar in spirit to model cards \cite{mitchell2019model} and model info sheets \cite{kapoor2023leakage}. These summaries demonstrate the value of structured machine learning workflow descriptions with consistent terminology. 

\paragraph{Machine learning design.}
The \textit{design} stage establishes the foundations of the workflow by connecting a scientific climate modeling or data analysis question to a machine learning task. The goal is to formulate the problem in a way that is scientifically meaningful and computationally tractable. This phase includes identifying and preparing appropriate training data, defining learning objectives, and selecting a class/space of candidate models that reflect prior knowledge, theory constraints, and application goals.

\textit{Scientific problem translation:} Identify the target process or climate modeling question and articulate modeling or statistical inference goals. Define constraints, required outputs, and relevance to decision-making or theory development.  For example, in Case Study 1c (Equation Discovery), on data-driven equation discovery, machine learning is used to address the challenges of developing better parameterizations for processes that occur at scales smaller than the model grid. Traditional parameterizations have relied on physical intuition, empirical fitting, and manual tuning, an approach that is labor-intensive, difficult to generalize, and prone to introducing biases. Machine learning offers a more systematic, data-driven strategy, using high-resolution simulations to learn functional relationships between resolved variables and subgrid effects. Zanna and Bolton (2020)~\cite{laureeqn} translated this scientific problem into a machine learning task by framing the search for closure models as a sparse regression problem, where the algorithm identifies functional relationships between grid-level variables and meaningful physically meaningful composite terms. This enables the discovery of interpretable equations that can refine parameterizations, improve model fidelity, and advance theory development.

\begin{figure}[h]
	\centering
	\includegraphics[width=0.65\textwidth]{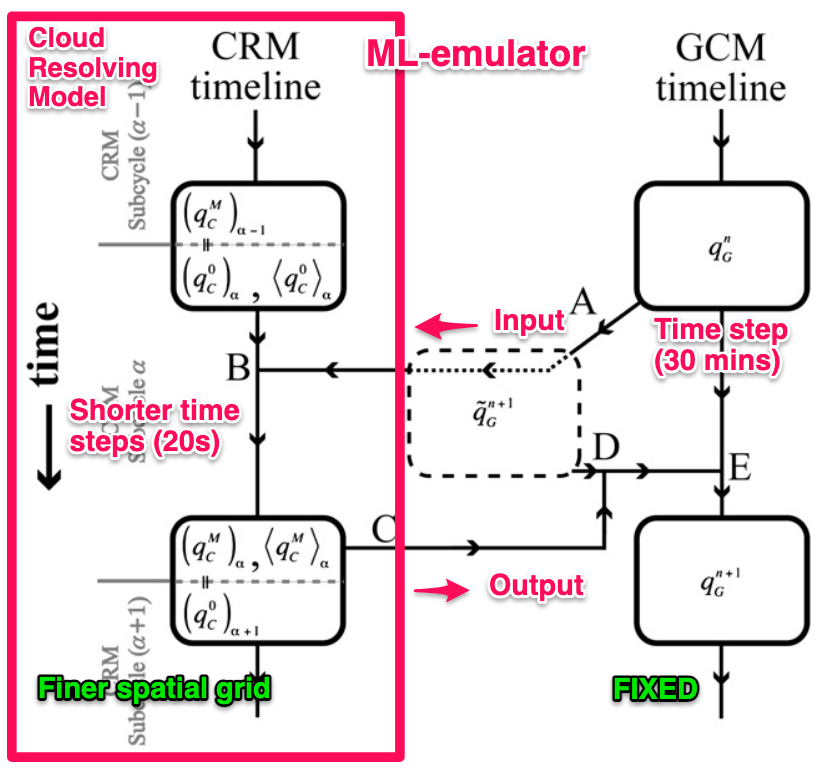}
	\caption{Data generating process for the Climsim dataset (annotated Figure 1 from \cite{benedict_structure_2009}). The Climsim data set aims to support the development of ML emulators for Cloud Resolving Model (CRM) that could be readily integrated into operational climate simulators such as the General Circulation Model (GCM).
		A.) Begin with macro-state of the entire grid cell, $q_{\,G}^n$ for tracked variables.
		B.) $q_G$ updates the internal convective state of the cloud model and supplies boundary conditions.
		C.) The cloud model steps forward in time, matching the 20 minute interval of the simulation.
		D.) A coarsened average of cloud internals yields output $\langle q_C^M \rangle_\alpha$ with resolution matching the GCM. Atmospheric states before and after cloud resolution are saved as input-output vectors for the dataset. 
		E.) Other processes and parameterizations are applied, cumulatively creating the next GCM macro state $q_{\,G}^{n+1}$. 
	}
	\label{fig:climsim-main}
\end{figure}

\begin{figure}[h]
	\centering
	\includegraphics[width=\linewidth]{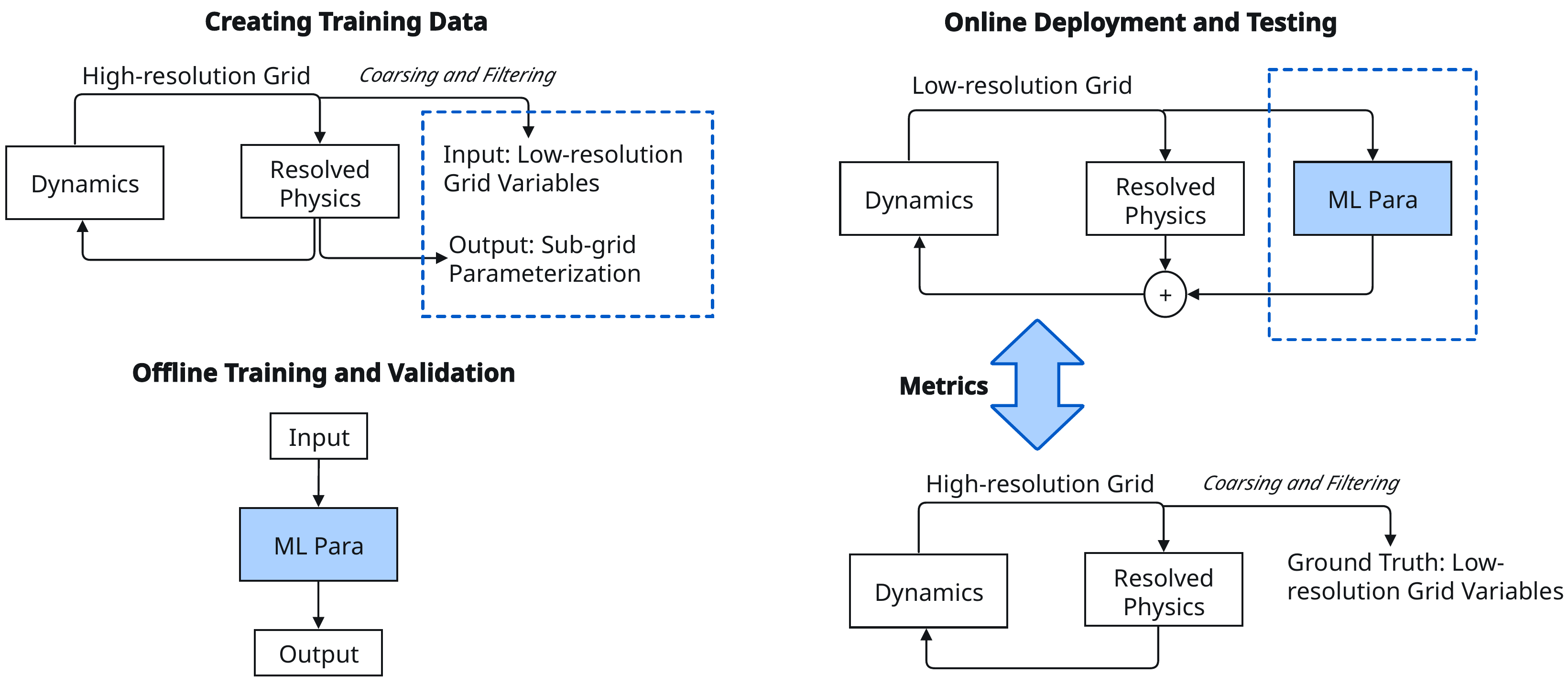}
	\caption{Workflow for machine learning-based parameterization of subgrid processes.}
	\label{fig:MLPara-main}
\end{figure}

\textit{Training data preparation:} Choose datasets (observational, reanalysis, or simulated) that are relevant to the task, taking into account not only representativeness, resolution, uncertainty, and coverage, but also the specific training and deployment needs implied by the chosen model architecture. The ClimSim \cite{yu2023climsim} (Case Study 1b) dataset was created specifically to support the development of hybrid models based on machine learning. Its design choices reflect a dual goal: to provide a machine-learning-ready input-output pair format while ensuring a straightforward downstream deployment into the host model. Inputs and outputs were carefully isolated to make the training task clear, while still preserving their "connections" with the hosting Earth system model (see Figure~\ref{fig:climsim-main}). 
Ross et al. (2023) \cite{zanna} (Case Study 1d: Benchmarking ML Parameterization Workflows) evaluated the different choices in the preparation of low-resolution data for training machine learning parameterization using high-fidelity simulations (see Figure~\ref{fig:MLPara-main}). In particular, there is no unique way of coarse-graining fields or compute subgrid tendencies, when constructing training inputs and outputs. By systematically benchmarking across options, Ross et al. (2023) \cite{zanna} demonstrated that machine learning performance can be highly sensitive to these early design choices. 
In Case Study 2a (Uncertainty Quantification for Remote Sensing), machine learning enabled simulations were used to create synthetic datasets for retrieval problems, allowing a rigorous quantification of uncertainties. Rather than relying solely on observational data, the workflow generated paired "true" states and retrievals, allowing the learning of error distributions. 
To overcome the sparsity of in-situ pCO$_2$ observations, Kim et al.\ (2024) \cite{kim_spatiotemporal_2024} (Case Study 3b: Transfer Learning for pCO$_2$ Upscaling) used simulated full-field data from ocean biogeochemical models to pre-train machine learning models before fine-tuning with sparse observational data. This two-step data preparation strategy leveraged the strengths of both simulated and observed datasets: coverage from models and realism from measurements. Such carefully designed sequencing of datasets enables machine learning to address fundamental observational gaps in climate science. 
Nathaniel et al.\ (2024) \cite{nathaniel2024chaosbench} (Case Study 3a: ChaosBench Benchmarking for Subseasonal Forecasting) illustrates how training data preparation and model space selection are tightly coupled. The scientific problem focuses on subseasonal-to-seasonal forecasting, i.e., predicting the future state of the climate system at lead times ranging from days to several weeks. Preparing training data for model development is closely tied to the forecasting strategy and model architecture. In the autoregressive setting, data can be prepared efficiently for stepwise prediction, but models must handle sequential roll-out and error accumulation. In the direct setting, by contrast, each forecast horizon requires its own training pairs, making data preparation increasingly costly as the horizon grows. 

\textit{Objective function specification:} Define learning objectives, loss functions, and metrics that reflect both statistical and scientific priorities (e.g., accuracy, physical consistency, uncertainty awareness). Case Study 1a (Neural Operators) provides an overview of training neural operators as surrogate models for climate processes. In this setting, training was framed as learning an operator rather than a pointwise predictor, with the objective extending beyond mean-squared error to include physics-informed residual terms that enforced PDE structure. In Case Study 1c (Equation Discovery), sparsity penalties were incorporated into the loss function to guide the model toward compact, interpretable equations rather than overfitting high-dimensional regressors. 

One open challenge for machine learning training in climate modeling is that there is often a disconnect between the loss functions used during training and the metrics we ultimately care about during deployment, such as distributional or spectral similarity, long-term stability or physical consistency (e.g., Case Studies 1c, 1d, 3a). These performance measures are typically not
straightforward to optimize with gradient descent. This mismatch can lead to models that perform well numerically in validation experiments but fail operationally. One promising recent strategy in machine learning could offer a potential solution where diffusion models are trained to be used as differentiable data priors, embedding scientific or empirical constraints directly into the objective function via generative modeling \cite{wu2024reconfusion}. 

\textit{Model space selection:} Specify model choices based on the scientific task, data availability, and structural constraints, including the selection of architecture families, inductive biases, and physics- or data-informed constraints. Among the case studies, Case Study 1a (Neural Operators) highlighted neural operators as surrogate models chosen specifically for their ability to approximate physical processes governed by differential equations. Unlike pointwise predictors, these operator learners were selected because they capture continuous spatiotemporal dynamics and encode conservation properties, aligning the model architecture with the underlying physics. In Case Study 2a (Uncertainty Quantification for Remote Sennsing), models were selected to support simulation-based inference and to explicitly represent uncertainty, showing how architecture choice can be guided by inference objectives rather than predictive accuracy alone. Case Study 2b (Inverse Parameter Inference) addressed an inverse problem: inferring physical parameters of a dynamical system from observed data distributions. Conditional normalizing flows were adopted here because they provide invertible mappings between parameters and distributions, enabling likelihood-free inference in complex dynamical settings. Finally, as discussed earlier, Case Study 3a (ChaosBench Benchmarking for Subseasonal Forecasting) compared architectures for subseasonal-to-seasonal forecasting, where model choice depends on both data availability and forecasting strategy: autoregressive surrogates for stepwise predictions versus direct predictors for longer lead times.

\paragraph{Machine learning development.}
In the \textit{development} stage, models are instantiated, trained, and systematically evaluated. The goal is to optimize among candidates in the \textit{model space} guided by the objective functions and training data. This stage often entails an iterative process of algorithmic refinement and scientific validation. This stage relies on controlled experiments and diagnostic tools to build confidence in model behavior and assess generalization capacity under scientific and statistical criteria. Most of the case studies we reviewed followed standard machine learning practices at this stage, which was made possible by careful problem formulation and data preparation in earlier stages of the workflow.

\textit{Validation Experiments:} Design experiments to evaluate model generalization and performance. Use diagnostic plots, predictive scores, cross-validation, and domain-specific metrics. In Case Study 1d (Benchmarking ML Parameterization Workflows), models were first tested offline against high-resolution simulations, with evaluation based on multiple metrics that captured both statistical accuracy and physical realism.

\textit{Modeling Algorithm Selection:} Choose training algorithms, optimization strategies, and regularization schemes. Tailor to model architecture and data characteristics. In Case Study 3b (Transfer Learning for pCO$_2$ Upscaling), the algorithmic design centered on transfer learning, where models were first pretrained on simulated full-field data and then fine-tuned on sparse observations to improve generalization across under-sampled regions.

\textit{Iterative Refinement:} Use insights from validation to adjust objective functions, training procedures, or data preprocessing. Incorporate scientific feedback into model revisions. Case Study 1d again illustrates this principle, showing the value of systematically comparing design choices on model architectures, feature sets, and definitions of subgrid forcing. Such a comparison yielded insights not just about predictive accuracy but also about which configurations were scientifically meaningful. This aligns with the Veridical Data Science framework \cite{Yu_Kumbier_2020}, where predictability, reproducibility, and stability are emphasized as essential for ensuring that workflow refinements lead to trustworthy scientific conclusions.

\paragraph{Machine learning deployment and Evaluation.}
The deployment and evaluation stage is especially critical in climate modeling, where machine learning components are integrated into larger Earth system models. Beyond offline testing using validation data, the key challenge is to evaluate online performance, ensuring that the machine learning surrogate interacts stably with other components and does not introduce long-term biases or instabilities. The goal is to guarantee that the model operates reliably in production environments and yields scientifically meaningful outputs. This stage also supports ongoing interpretation, uncertainty assessment, and the feedback of new insights to earlier design stages, thereby completing the workflow loop.

\textit{Integration and Production:} Insert the trained model into a larger climate modeling pipeline (e.g., via emulators or closures), policy system, or forecasting workflow. Ensure computational feasibility and traceability. Case Study 1b (ClimSim) exemplifies this stage: models trained on the ClimSim dataset were deployed directly into DOE’s E3SM model using containerized workflows and PyTorch–Fortran bindings. This setup made it possible to evaluate ML parameterizations in the same production environment as traditional physical parameterizations, ensuring traceability, stability, and reproducibility of the coupled system.

\textit{Evaluation and Interpretation:} Conduct ongoing testing under new conditions or climate regimes. Assess uncertainty, diagnose failure modes, and interpret model outputs in scientific or operational contexts. In Case Study 1c (Equation Discovery for Ocean Mesoscale Closures), evaluation required testing whether discovered equations preserved physical realism and stability over long online simulations, beyond fitting offline statistics. In Case Study 1d (Benchmarking ML Parameterization Workflows), evaluation emphasized systematic comparison of models under multiple metrics, including physically meaningful diagnostics such as energy spectra and decorrelation times, not just standard error measures. In Case Study 2a (Uncertainty Quantification for Remote Sensing), interpretation focused on whether ML-enabled simulators provided calibrated and transferable uncertainty estimates for retrieved geophysical variables under new observational conditions. For scientific machine learning applications like climate modeling, meaningful evaluation extends beyond predictive accuracy to include considerations of scientific validity, interpretability, and robustness.

\textit{Informing Upstream Design:} Use post-deployment insight to revisit earlier design decisions. Inform refinement of training data, objective function, or model choice. Case Study 3a (ChaosBench Benchmarking for Subseasonal Forecasting) illustrates this principle: benchmarking across architectures and strategies revealed systematic trade-offs (e.g., autoregressive vs. direct models) that in turn suggested new directions for data preparation, loss design, and hybrid architecture choices. 


\section{Conclusion and Discussion}
This paper has illustrated how thoughtful design choices can shape rigorous and interpretable machine learning workflows in climate modeling. Through eight case studies, we highlight strategies for applying machine learning to improve prediction, inference, and calibration across a range of climate science tasks. Rather than treating machine learning as a black-box optimization tool, each workflow exemplifies an integrative approach that connects empirical modeling with simulation, theory, and constraints.

We aim to show that robust applications of machine learning in climate science are less about identifying the "best" model in an absolute sense, and more about building workflows that support uncertainty quantification, reproducibility, and scientific interpretation. As machine learning becomes increasingly embedded in the climate modeling literature, sustained attention to workflow design will be essential for creating tools that are both effective and scientifically trustworthy.

Across the diverse case studies discussed, several recurring patterns emerged. Several workflows benefited from a modular design in their conceptual structure, which allows models, diagnostics, and inference components to be reused or swapped with minimal friction. Another common theme was the integration of physical relevance into the modeling process, whether through physics-informed architectures, conservation-based constraints, or domain-guided research designs. Finally, several case studies showed that offline skill is not an adequate indicator of online deployment performance. In climate modeling, it is the online setting, where models must interact with complex feedbacks over long simulations, that ultimately tests their stability, physical realism, and sustained accuracy.

Despite these advances, important challenges remain. \textit{Reproducibility} remains a concern in the climate modeling literature, partly due to inconsistent terminology, fragmented software tools, and ad hoc workflow documentation. Greater standardization, both in language and in modular computational practices, could help the community build more transparent and shareable workflows. \textit{Interpretability} also remains an open question. While some workflows benefit from inherently explainable models, many require more systematic experimentation to assess which components drive performance and how design choices influence scientific conclusions. Workflows in climate modeling could also place a greater emphasis on formally modeling uncertainty, not only to improve predictive reliability but also to support scientific interpretation and understanding. 


\textit{Scalability} is another pressing issue. As machine learning models are increasingly applied to ensemble simulations and long-term climate projections, the computational burden of training and inference grows rapidly. Efficient, workflow-aware strategies will be needed to scale machine learning methods without compromising physical fidelity. 

Finally, a key frontier lies in better integrating machine learning workflows with \textit{scientific reasoning} itself. In addition to being a tool for improving prediction, machine learning also holds potential for hypothesis generation, theory refinement, and mechanistic understanding. Workflows that systematically integrate physical insight, support diagnostic feedback, and remain interpretable to domain scientists will be central to the future of machine-learning-driven climate science.


\newpage
\setcounter{table}{0}
\renewcommand{\thetable}{S\arabic{table}}%
\setcounter{figure}{0}
\renewcommand{\thefigure}{S\arabic{figure}}%
\setcounter{section}{0}
\renewcommand{\thesection}{S\arabic{section}}%

{\Large \bf \noindent Supplement for ``Machine Learning Workflows in Climate Modeling: Design Patterns and Insights from Case Studies''}





\section{Physics-First Case studies}
\label{sec:physicsfirst}

\subsection*{Case Study 1a: Neural Operators for Physics-Relevant Emulation}

\paragraph{Scientific problem.} One of the earliest promises of machine learning for improving climate models is its ability to approximate computationally expensive physical processes governed by partial differential equations (such as atmospheric dynamics or ocean circulation) \cite{rasp2018deep}, replacing high-fidelity numerical solvers. Machine learning \textit{emulators}, which serve as \textit{surrogate models}, significantly reduce simulation cost while preserving key input–output behaviors, and support data assimilation, uncertainty quantification, or policy‐driven scenario analyses. 

\begin{figure}[h]
    \centering
    \includegraphics[width=\linewidth]{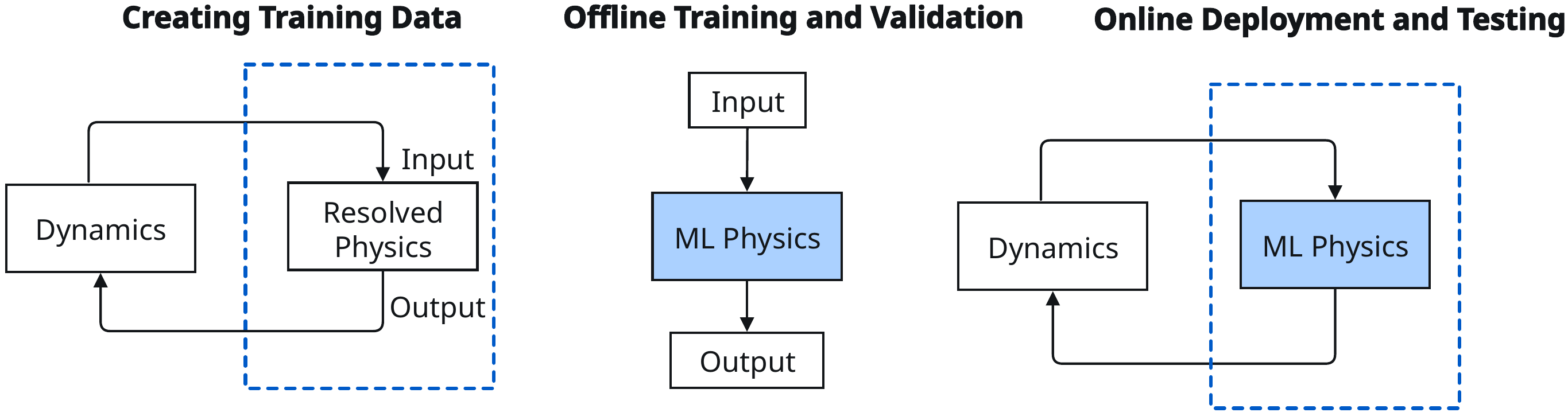}
    \caption{Workflow for training and deploying ML-based surrogate models.}
    \label{fig:surrogate}
\end{figure}

\paragraph{Machine learning design.}
Building a ML surrogate model for a climate process, or an \textit{ML climate emulator}, starts with the generation of training data from high-resolution simulations or resolved physics components within a larger climate model (Figure~\ref{fig:surrogate}, left). These simulations provide paired inputs and outputs. In the \textit{offline} training phase (Figure~\ref{fig:surrogate}, center), a machine learning model is trained to map these inputs to outputs. Once constructed, the surrogate model is \textit{deployed online} (Figure~\ref{fig:surrogate}, right), replacing the original component based on numerical solvers, and is integrated back into the broader dynamics model.

One popular \textit{model choice} for surrogate models of climate processes is the class of neural operators \cite{kovachki2023neural}, which focuses on learning spatiotemporal fields $u(x,t)$ defined over spherical or Cartesian domains with heterogeneous boundary and forcing conditions. The climate emulator can be formatted as the following operator learning problem:
\[
\mathcal{G}: \;\; u_0(\cdot)\;\mapsto\; u_T(\cdot),
\]
where $u_0$ is the initial condition and $u_T$ the solution at lead time $T$.  Equivalently, for time‐stepping, $\mathcal{G}_\Delta: u_{t}\mapsto u_{t+\Delta}$. This is a regression task in an infinite-dimensional space, as inputs and outputs are both functions (fields). Neural operator layers are typically applied in an iterative fashion. For instance, in the FNO formulation each update takes the form (cf.\ eqn.~(2) in~\cite{li2020fourier})
\[
  v_{t+1}(x) \;=\; \sigma\Bigl(W\,v_t(x)\;+\;(\mathcal{K}(a;\phi)\,v_t)(x)\Bigr),
  \quad\forall x\in D,
\]
where $\mathcal{K}(a;\phi)$ is the kernel integral operator parameterized by $\phi\in\Theta_{\mathcal K}$ (with $a\in\mathcal A$ the input coefficients), given $v_t$ is the feature field at iteration $t$, $x\in D$ a spatial location, $W$ a learned linear map,  and $\sigma$ a component-wise activation function. Different neural operator variants differ by their approaches to setting up the kernel integral operator, $\mathcal{K}(a;\phi)$, which share a generic form, 
\[
  (\mathcal{K}_\phi\,v)(x) \;=\;\mathcal{T}^{-1}\!\bigl(R_\phi \cdot (\mathcal{T}\,v)\bigr)(x).
\]
Here, $\mathcal{T}$ is a linear transform, $R_\phi$ a learned multiplier (or kernel), and $\mathcal{T}^{-1}$ its inverse.  Different choices of $\mathcal{T}$ yield different operator architectures:
\begin{itemize}
  \item \textit{Fourier Neural Operator (FNO):} $\mathcal{T}=\mathcal{F}$ (Fast Fourier Transform); this variant excels at learning homogeneous, periodic dynamics such as idealized fluid flows.  ~\cite{li2020fourier}
  \item \textit{Spherical FNO (SFNO):} $\mathcal{T}= \mathcal{F}_{SHT}$ (Spherical Harmonic Transform); this architecture is designed for global weather and climate fields on the sphere with built‐in rotational equivariance. ~\cite{bonev2023spherical}
   \item \textit{Wavelet Neural Operator (WNO):} $\mathcal{T}=\mathcal{W}_{\mathrm{db}m}$ (Daubechies Wavelet Transform); this variant leverages the time-frequency localization of Daubechies wavelets to capture fine-scale spatial and spectral features, such as building a digital twin for Earth’s air temperature forecasting. ~\cite{tripura2022wavelet}
    \item \textit{Graph Neural Operator (GNO):} instead of a fixed spectral basis, GNO constructs a spatial proximity graph over the discretization and approximates the integral operator via message‐passing updates with learned kernels; this mesh‐independent approach handles unstructured grids (e.g.\ ocean models, river networks) through Nyström‐style subsampling and local aggregation. ~\cite{li2020graph}
  \item \textit{Geometry‐Informed Neural Operators (GINO):} embeds transforms derived from local chart maps or differential operators on manifolds. This model can be used for flows over complex terrains or ice‐sheet dynamics. \cite{li2023geometryinformed}
\end{itemize}
Modularity of the above setup means any $\mathcal{T}$ can be swapped in to inject prior knowledge (spectral, geometric, multiscale) while keeping the same end‐to‐end training pipeline.

The \textit{training data} of neural operators come from high‐resolution models or observational ``truth'' trajectories. In Li et al.’s FNO experiments, the model was trained on synthetically generated PDE solutions, specifically, two‐dimensional Burgers’, Darcy flow, and Navier–Stokes snapshots on periodic domains~\cite{li2020fourier}. For the SFNO case, Bonev et al. used simulated atmospheric‐dynamics data, including shallow‐water‐equation roll‐outs and reanalysis‐like surface wind and geopotential height fields on a spherical grid~\cite{bonev2023spherical}. Prior to training, all fields are interpolated to a common resolution (with up/down‐sampling as needed), temporal cadences are synchronized to a uniform $\Delta t$, and systematic biases are corrected via ensemble‐mean subtraction.

The canonical loss for neural operators can be written as
\[
  \mathcal{L}(\theta)
  = 
  \underbrace{\bigl\|\mathcal{G}_\theta(u_0)-u_T\bigr\|_{L^2}^2}_{\text{data loss}}
\]

In practice, FNO \cite{li2020fourier} and SFNO \cite{bonev2023spherical} both minimize this $L_2$ term. However there are alternative approaches, such as the Physics-Informed Neural Operator (PINO) by Li et al. (2021) \cite{li2023physics}, which minimizes the following combined loss

\[
  \mathcal{L}(\theta)
  = 
  \underbrace{\bigl\|\mathcal{G}_\theta(u_0)-u_T\bigr\|_{L^2}^2}_{\text{data loss}}
  \;+\;
  \underbrace{\bigl\|\mathcal{R}\{\mathcal{G}_\theta(u_0)\}\bigr\|_{L^2}^2}_{\text{PDE residual loss}},
\]

where \(\mathcal{R}[\cdot]\) denotes the PDE residual operator. The PINO loss uses coarse-resolution data for the first term and enforces the PDE residual at higher resolution, without introducing an explicit weighting parameter between them.

\paragraph{Model development.}
Given the modularized model design, neural operators are trained using standard deep learning practices, given carefully prepared training data and an appropriate objective function.

\paragraph{Model deployment and evaluation}
Once trained, neural operators are often deployed to generate rollouts of the PDEs they emulate, either as standalone surrogates or integrated within larger simulation models. They are evaluated based on the accuracy and stability of the resulting trajectories, with metrics that assess conserved quantities, long-term behavior, and physical consistency. In \cite{li2020fourier}, the FNO was assessed primarily by its relative \(L^2\) error on held‐out PDE solutions (including Darcy flow, two‐dimensional Burgers’ and Navier–Stokes) alongside inference latency. 
In \cite{bonev2023spherical}, the SFNO was evaluated using RMSE on global 500 hPa geopotential‐height and 10 m surface‐wind forecasts, spectral‐energy fidelity, and long‐term stability over extended rollouts. 
%
%

\subsection*{Case Study 1b: Training Data and Development Environment for High-Resolution Physics Emulators in Hybrid Multi-Scale Climate Simulators}
\paragraph{Scientific problem.} This case study addresses a similar scientific problem as Case Study 1a: improving the representation of subgrid-scale atmospheric processes in climate models. It focuses on the challenge of developing stable and accurate hybrid machine learning–physics climate simulations by introducing a comprehensive dataset, {\em ClimSim} \cite{yu2023climsim}, and an end-to-end training and deployment framework, {\em ClimSim-Online} \cite{yu2024climsim}, for ML-based surrogate models. The {\em ClimSim} dataset is a large-scale, publicly available resource designed to support the development of machine learning parameterizations for processes such as cloud formation and extreme rainfall, which are \textit{unresolved} in coarser-resolution models. These processes are typically represented using parameterizations. A computationally more expensive yet better alternative is high-resolution cloud-resolving models (CRMs) \cite{khairoutdinov2005simulations}, which explicitly simulate convection and cloud dynamics at grid scales of a few kilometers, providing more physically accurate representations. 

\paragraph{Machine learning design.}
\textit{ClimSim} \citep{yu2023climsim} uses cloud-resolving model (CRM) simulations to generate high-resolution training data, enabling surrogate models that can replicate CRM-level outputs at a fraction of the computational cost. These surrogates are expected to replace traditional parameterizations in Earth system models (ESMs), offering a more data-driven and scalable alternative. A central design goal of \textit{ClimSim} is to provide a clean and modular interface between physical simulations and machine learning workflows. To do this, data are generated from operational runs of the Department of Energy E3SM Multiscale Modeling Framework (E3SM-MMF) \citep{e3sm2023}, where a CRM is embedded in each grid cell of a host climate model \citep{khairoutdinov2005simulations}, a configuration known as \textit{superparameterization} (Figure~\ref{fig:climsim-main} in the main text). This multiscale architecture naturally separates large-scale resolved dynamics from the fine-scale physics to be emulated, making it well suited for supervised machine learning training.

The resulting dataset consists of ten years of globally distributed input–output pairs at approximately 1.5$^\circ$ horizontal resolution and 20-minute temporal intervals. Inputs include large-scale atmospheric state variables and boundary conditions (e.g., temperature, wind, solar insolation, and surface fluxes). At the same time, outputs reflect the CRM-resolved effects of subgrid processes, such as cloud formation, turbulence, radiation, and precipitation, which are readily inputs for the hosting Earth system model.

Per its design, the ClimSim data set invites a wide range of high-dimensional regression machine learning models: $f: X \rightarrow Y$, which accurately map resolved high-dimensional atmospheric inputs $X$ to subgrid high-dimensional physical tendencies $Y$. Once trained, such a model can replace the CRM in online simulations. ClimSim was offered as \href{https://www.kaggle.com/competitions/leap-atmospheric-physics-ai-climsim}{a Kaggle Competition} in 2024, where the mean squared error (MSE) was used as the primary loss function due to its simplicity, scalability, and ease of comparison between submissions. Although more physically informed losses may be desirable, in principle, they were not easy to implement for such an online competition.

\paragraph{Model development.}

Given the thoughtful and well-structured design of the ClimSim dataset, model development largely follows standard best practices in supervised machine learning, including architecture selection, loss minimization, regularization, and validation on held-out samples.

\paragraph{Model deployment and evaluation}
Ultimately, the goal of the ClimSim project is to enable the development of machine learning emulators that can be stably integrated into the E3SM climate model for long-term simulations. In the ``online'' testing setting, the trained machine learning model replaces the CRM and provides the predicted sub-grid tendencies used to advance the atmospheric state in time. Unlike offline evaluation where model accuracy is assessed in independent samples using standard metrics such as weighted RMSE, online performance critically depends on the model's ability to maintain physical stability and coherence over extended rollouts. Even small numerical inconsistencies can accumulate, leading to divergence from physically realistic behavior. Offline accuracy alone does not guarantee stable coupling in the full system.

To facilitate reproducible and efficient online testing, {\em ClimSim-Online} provides a containerized evaluation framework \cite{boettiger2015introduction} that enables machine learning models to be deployed directly within the Fortran-based E3SM model. Cross-language integration is handled via PyTorch–Fortran bindings using TorchScript or FTorch \cite{atkinson2025ftorch}, allowing models trained in Python to be inserted into E3SM with minimal modification. This workflow greatly simplifies the process of testing hybrid ML–physics models in a production climate modeling environment, accelerating the development of robust ML parameterizations.

\subsection*{Case Study 1c: Data-Driven Equation Discovery in Climate Modeling}

\paragraph{Scientific problem.} As previously discussed, climate models are built to approximate the behavior of Earth's systems by numerically solving the physical equations that govern their dynamics. These equations are evaluated on a grid, where state variables are updated over time using discretized forms of the governing equations. Processes that occur at scales larger than the grid resolution are considered \textit{resolved}. Processes that are smaller than the grid resolution, such as convection, cloud formation, or turbulence, are called \textit{unresolved} and require a \textit{closure}, a strategy to complete the system of equations by adding an approximation term that represents the influence of unresolved processes on the resolved dynamics to approximate their impact on the resolved system. The most common form of closure is a \textit{parameterization}, which is based on simplified, often empirical, formulas to represent the average effect of the \textit{sub-grid} processes. Parameterizations are traditionally developed through a mix of physical intuition, empirical fitting, and manual tuning. This process is often labor-intensive, difficult to generalize across regimes, and introduces biases and uncertainty into the system for which it is meant to improve. As climate models become more complex, there is a growing need for more systematic, data-driven strategies to identify and refine these parameterizations.

For example, modern ocean models run on grids that are too coarse to explicitly resolve mesoscale eddies, i.e., the coherent vortices that organize otherwise chaotic ocean turbulence. Parameterization is needed to ``close" the missing physics by modeling subgrid behaviors~\cite{laureeqn}.

\paragraph{Machine learning design.} Equation discovery through machine learning is an active research area, where physically meaningful and interpretable mathematical expressions are learned in a data-driven fashion to describe the effects of unresolved processes. It can also lead to the generation of scientific hypotheses and the refinement of theories in modeling physical systems ~\cite{brunton2016discovering, champion2019data}. 
Zanna and Bolton (2020)~\cite{laureeqn} applied equation discovery to parameterize the impact of mesoscale eddies on large-scale circulation in ocean models. The goal is to learn an interpretable closure model 
\[
  \tau = \mathcal{F}\bigl(u,\,\nabla u,\,\ldots\bigr)
\]
for the subgrid stress \(\tau\) as an explicit function of resolved velocity fields \(u\). The \textit{training data} were based on high-resolution simulations from the MITgcm model \cite{marshall1997finite} (Figure~\ref{fig:MLPara-main} in main text, top-left). After long spin-up periods, 1,000 time slices of model output were saved. A filter was applied to produce coarse-grained data. Subgrid eddy forcing terms were computed by comparing high-resolution and coarse-grained fields, which serve as the target outputs for training. The training inputs consist of coarse-resolution velocity and temperature fields and their spatial derivatives, including physically meaningful composite terms like vorticity and divergence, which are \textit{candidate} predictors for the subgrid eddy forcing in the \textit{equation discovery} framework.

The \textit{model choice} for equation discovery in~\cite{laureeqn} formulates the discovery process as a sparse regression problem by employing a Relevance Vector Machine, or RVM, \cite{tipping2001} to identify a minimal subset of basis functions that best reproduce the subgrid forcing. They then benchmarked this RVM-derived closure against two alternatives: (1) a fully connected physics-constrained neural network (FCNN) trained on the same data and (2) the parameterization of AZ17: an analytically derived, deformation-based eddy Reynolds stress model \cite{lz14}.

\paragraph{Model development.} During model training, validation was performed offline using held-out time slices and assessed through statistical comparisons between predicted and true subgrid forcings, including correlation, spatial statistics, and higher-order moments.


\paragraph{Model deployment and evaluation.} The discovered RVM closure, once deployed \textit{online} and embedded into the ocean‐general circulation models, is evaluated over long simulations (Figure~\ref{fig:MLPara-main} in main text, right). Its performance was assessed by comparing the resulting large-scale flow structures and energy distributions to high-resolution reference runs, in terms of both stability and physical realism. This is compared against the physics‐constrained FCNN, trained on the same filtered MITgcm output, which delivers the most robust and accurate reproduction of mean kinetic energy, spectral backscatter, and extreme‐event statistics in prognostic runs, and the AZ17 deformation‐based parameterization \cite{lz14}, which yields a compact, interpretable formula and matches offline statistics closely but requires similar attenuation/tuning to remain stable when coupled online. 

These comparisons illustrate the trade‐off between interpretability and ease of integration (RVM, AZ17) versus prognostic stability and energetic fidelity (FCNN). Through the RVM pipeline, Zanna and Bolton (2020)~\cite{laureeqn} discovered equations that recovered known advection–diffusion forms with only a handful of terms, yielding interpretable closure models whose performance matched or exceeded black‐box deep learning models, while offering the advantage of direct physical insight and easier integration into existing codebases of Earth system models. Such discovered equations can be further interpreted against known physical laws and refined using domain knowledge~\cite{cranmer2020discovering}.

\subsection*{Case Study 1d: Benchmarking Machine Learning Parameterization Workflows for an Idealized Model}

\paragraph{Scientific problem.}
This case study addresses the same scientific problem of improving subgrid parameterization as described in the previous case study. Ross et al.\ (2023) \cite{zanna} conducted a systematic benchmarking study of machine learning parameterization workflows using a two-layer quasi-geostrophic (QG) channel model, a simplified yet dynamically rich system often used to study mesoscale eddies. This model preserves key physical features such as baroclinic instability and the inverse energy cascade, which are essential to geophysical turbulence, while remaining computationally efficient enough to support large ensemble experiments.

\paragraph{Machine learning design.}
For each model layer $m\in\{1,2\}$, the simulation updates the potential vorticity (PV) field 
$\hat q_m$ in spectral space according to:
\begin{equation}
\frac{\partial \hat q_m}{\partial t}
      \;=\;
      \mathcal{R}_m
      \;+\;
      \hat S_t.
\label{eq:qg_coarse}
\end{equation}
Here $\mathcal{R}_m$ includes the \textit{resolved} physical terms, such as advection, mean shear, and dissipation, that are directly computed by the numerical solver. The term $\hat S_t$ represents the \emph{subgrid forcing}, which accounts for the effects of unresolved small-scale dynamics that are not captured on the grid. $\hat S_t$ is ignorable for higher-resolution grids and shorter emulation, but could lead to substantial biases for a coarser grid if not accounted for correctly. The \textit{subgrid forcing} plays the same role as a parameterization, as it approximates the influence of finer-scale processes on the resolved dynamics. To enhance the reliability of low-cost simulations on coarser-resolution grids, machine learning parameterizations, $f_\theta(x_t)$, are to be trained to predict $\hat S_t$ given coarse-resolution inputs $x_t$, i.e., 
\[
f_\theta:\ \mathbf{x_t}\;\longmapsto\;\hat S_t.
\]
Given $\hat S_t$, the system (Equation (\ref{eq:qg_coarse})) can then be solved pseudospectrally, using real-space computations for nonlinear terms and spectral-space integration via a third-order Adams–Bashforth scheme [ref]~\citep{Abernathey2015}.

\textit{Training data} for building the machine learning parameterizations are derived from simulated high-resolution fields (Figure~\ref{fig:MLPara}, top-left). For example, in Ross et al.\ (2023) \cite{zanna}, the parameterizations were trained to improve emulation on a $64\times64$ grid while the training data were derived from simulated \textit{field snapshots} on a $256 \times 256$ grid. 
%
$250$ independent runs were generated on the high-resolution grid, where the systems (Equation~\ref{eq:qg_coarse}) are solved forward from random noise for 10 simulation years, reaching a quasi-steady state after a spin-up of ~3-5 years (depending on the initial condition). The systems are sampled every 1,000 hours so that successive snapshots are weakly correlated. 
For each snapshot from the high-resolution grid, a given filter \& coarse-graining operator $\mathcal{A}$ is applied to derive a coarse-grid \textit{input} $x$ and a \textit{label} $y$ is calculated using a given subgrid forcing definition $S$. In Ross et al.\ (2023) \cite{zanna}, three different choices for $\mathcal{A}$, and five different choices for $S$ were considered as part of the benchmarking effort to estimate the impacts of these design choices. Seven possible input features were included in $x$. 

For the machine learning \textit{model choice}, Ross et al.\ (2023) \cite{zanna} focused on \textit{physics-sonstrained fully-connected neural network (FCNN)}, an eight-layer fully-connected design as in \citep{GuillauminZanna2021}. For every combination of coarse-graining filter $\mathcal{A}$ and subgrid forcing definition $S$, a different FCNN was trained, minimizing the mean-squared error (MSE). 

\paragraph{Model development.}
During model development, circular padding was used to respect the periodic boundaries of the simulation domain. To prevent spurious growth of potential vorticity (PV, $q$) across the domain and to maintain numerical stability (consistent with Courant–Friedrichs–Lewy conditions), a constraint was applied to the final layer of the neural network: its output must have zero spatial average when predicting PV forcing or momentum forcing. This ensures that the net effect of the predicted forcing does not introduce unwanted imbalances in the systems. During the \textit{offline} model validation, the predicted $\hat S$ is compared with the ground truth $S$ on held-out snapshots using \textit{coefficient of determination} and \textit{Pearson correlation}.

\paragraph{Model deployment and evaluation.} Once trained and validated \textit{offline}, each ML parameterization $\hat S$ is embedded in the prognostic two–layer QG solver at low-resolution as the subgrid term in \eqref{eq:qg_coarse}. The low-resolution model is then integrated forward to evaluate \emph{online} fidelity and stability as described below. 

\begin{enumerate}
    \item \textbf{Pointwise Difference of Temporally-Averaged Power Spectra and Fluxes:} 
\begin{equation}
\label{eq:rmse-spectra}
\mathrm{RMSE}_{\mathrm{spec}}(\mathrm{sim}_1,\mathrm{sim}_2;\,f)
\;=\;
\sqrt{\;
   \frac{1}{|\mathcal K|}
   \sum_{k\in\mathcal K}
     \bigl(
       f_{\mathrm{sim}_1}(k)-f_{\mathrm{sim}_2}(k)
     \bigr)^{2}
   }\!,
\end{equation}
where $\mathcal K$ is a fixed set of isotropic wavenumbers common to both simulations. This metric provides information on long-term, statistical differences.
%
\item \textbf{Distributional Difference of Spatially-Flattened Probability Distributions:}
\begin{equation}
\label{eq:distrib-diff}
W_1(\mathrm{sim}_1,\mathrm{sim}_2)
\;\equiv\;
\int_{-\infty}^{\infty}
\bigl|
  F_{\mathrm{sim}_1}(f;z)\;-\;F_{\mathrm{sim}_2}(f;z)
\bigr|\,\mathrm d z,
\end{equation}
where $W_1$ is the Wasserstein-1 distance, $F$ is the cumulative distribution function of a given quantity $f$. 
\item \textbf{Decorrelation-Time Difference:} 
\begin{subequations}
\begin{align}
\mathrm{decorr\_time}(\mathrm{sim}_1,\mathrm{sim}_2)
&=
\mathbb E_{q_0,\epsilon}\!
  \Bigl[
    \min_{t}\bigl\{
      t:\;
      \mathrm{Corr}\bigl(
        q^{(t)}_{\mathrm{sim}_1}(q_0),
        q^{(t)}_{\mathrm{sim}_2}(q_0+\epsilon)
      \bigr)\le\delta
    \bigr\}
  \Bigr],
\label{eq:decorr-time} \\[4pt]
\mathrm{decorr\_diff}(\mathrm{sim}_1,\mathrm{sim}_2)
&=
  \mathrm{decorr\_time}(\mathrm{sim}_1,\mathrm{sim}_1)
 -
  \mathrm{decorr\_time}(\mathrm{sim}_1,\mathrm{sim}_2).
\label{eq:decorr-diff}
\end{align}
\end{subequations}
Here $q^{(t)}_{\mathrm{sim}}(q_0)$ is the PV snapshot after time $t$ when the simulation starts from $q_0$; $\epsilon$ is an i.i.d.\ Gaussian perturbation with $\sigma=10^{-10}$; $\delta$ is fixed at 0.5; and Corr refers to Pearson correlation. Expectations are approximated empirically over five random pairs $(q_0,\epsilon)$.
\end{enumerate}

Using these diagnostic distances in Equations.~\eqref{eq:rmse-spectra}–\eqref{eq:decorr-diff}, Ross et al.\ (2023) \cite{zanna} proposed \emph{similarity score} that measures how much closer a ML-parameterized low-resolution run (``param'') is to the high-resolution reference (``hr’’) than the unparameterized low-resolution baseline (``lr’’):
\begin{equation}
\label{eq:similarity}
\mathrm{Similarity}(\text{param},\text{hr};\mathrm{diff})
\;=\;
1\;-\;
\frac{\mathrm{diff}(\text{param},\text{hr})}
     {\mathrm{diff}(\text{lr},\text{hr})}\, .
\end{equation}
%
%
Similarity scores (Equation \eqref{eq:similarity}) were calcuated for each distance metric and averaged into a single online performance index that was used for comparison across design choices.

\section{Data-First Case Studies}
\label{sec:datafirst}
\subsection*{Case Study 2a: Simulation-Based Uncertainty Quantification for Remote Sensing}
\paragraph{Scientific Problem.} Remote sensing data from Earth-orbiting satellites provide one of the most comprehensive sources of global observational information for studying Earth’s climate. Since the launch of TIROS-1 (Television Infrared Observation Satellite-1) in the early 1960s, the remote sensing record has grown substantially. As of 2025, NASA holds approximately 120 petabytes of Earth observation data, a volume expected to grow to over 600 petabytes by 2030 based on planned satellite missions \cite{nasaearthdata}. These datasets serve multiple roles in climate science: they support machine learning applications by offering large-scale observational archives; they inform the development of subgrid-scale parameterizations in physical models; and they provide a reference against which climate model output can be validated and diagnosed.

However, it is often overlooked that remote sensing instruments do \textit{not} observe physical quantities directly. Instead, they measure electromagnetic spectra, which must be converted into estimates of physical variables, such as temperature, humidity, or aerosol concentration via inference procedures known as \textit{retrieval algorithms}. For these retrieved quantities to be scientifically useful in any of the aforementioned roles, they must be accompanied by uncertainty estimates that account for sources of uncertainty, such as instrument limitations and algorithmic assumptions.

\begin{figure}[h]
\centering
\tikzstyle{every node} = [align=center]
\begin{tikzpicture}[scale=0.8]
    \matrix [column sep=3mm,row sep=2mm] {
        \node(TrueState) {\color{gray}{True} \\ \color{gray}{state}}; &
        \node(ForwardFunction) {\color{gray}{Forward} \\ \color{gray}{function}}; &
         \node(NoiselessRadiance) {\color{gray}{Noiseless} \\ \color{gray}{radiance}}; & 
        \node(Instrument) {\color{gray}{Instru-} \\ \color{gray}{ment}}; &  
        \node(Observation) {Obser- \\ vation}; & 
        \node(Retrieval) {Retrieval \\ algorithm}; & 
        \node(StateEstimate) {State \\ estimate}; \\
        \node (X0) {\color{gray}{$X_{t}(\mathbf{s})$}};
        & \node[rectangle, thick,draw,gray] (F0) {\color{gray}{$\mathrm{F}(\cdot)$}};
        & \node (calY0) {\color{gray}{$Y_{0t}(\mathbf{s})$}}; 
        & \node[rectangle, thick,draw,gray] (calY0+eps) {\color{gray}{$Y_{0t}(\mathbf{s})E_{t}(\mathbf{s})$}};
        & \node (Y0) {$Y_{t}(\mathbf{s})$};
        & \node[rectangle, thick, draw] (R0) {$\widehat{\mathrm{R}}\left(\cdot, \widehat{\mathrm{F}}, \beta_{t}(\mathbf{s}), \Gamma_{t}(\mathbf{s}) \right)$};
        & \node (hatX0) {$\widehat{X}_{t}(\mathbf{s})$} ; \\
    }; 
    \draw[->, gray,thick] (X0) -- (F0);    
    \draw[->, gray,thick] (F0) -- (calY0);  
    \draw[->,gray,thick] (calY0) -- (calY0+eps);  
    \draw[->, gray,thick] (calY0+eps) -- (Y0); 
    \draw[->,thick] (Y0) -- (R0); 
    \draw[->,thick] (R0) -- (hatX0); 

    \node[below of = calY0+eps, xshift = 0, yshift = -5, gray] (joint) {$\mathrm{P}(\textcolor{gray}{\mathrm{X}}, \widehat{\mathrm{X}})$};
	\node[below of = joint, xshift = 0, yshift = -5, gray] (conditional) {$\mathrm{P}(\textcolor{gray}{\mathrm{X}} | \widehat{\mathrm{X}})$};
	\draw[->, thick,gray] (X0) |- (joint);
	\draw[->, thick,gray] (hatX0) |- (joint);
	\draw[->, thick,gray] (joint) -- (conditional);
\end{tikzpicture}
\caption{Conceptual diagram of a remote sensing observing system. Item in gray are unknown, and items in black are known. $t$ indicates time point, and $\mathbf{s}$ is a two-dimensional (or in some cases, three-dimensional) spatial index, e.g., $\mathbf{s}=(lat,lon)$. The key quantities of interest are the joint distribution of the state vector and its estimate, and ultimately the conditional distribution $\mathrm{P}(\textcolor{gray}{\mathrm{X}} | \widehat{\mathrm{X}})$.}
\label{fig:rs1}
\end{figure}
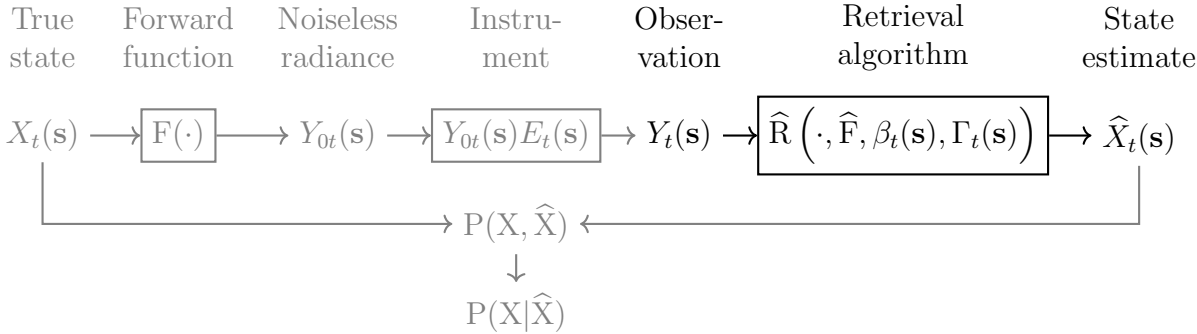

Figure~\ref{fig:rs1} is a conceptual diagram of a remote sensing observing system. At a given location $\mathbf{s}=(lat,lon)$ and time $t$, the true state of Earth system is represented by random vector $X_{t}(\mathbf{s})$ of dimension $d_{X}$. Nature (notionally represented by a {\em forward function}) converts $X_{t}(\mathbf{s})$ into a noise-free, continuous electromagnetic spectrum, $Y_{0t}(\mathbf{s})$ of dimension $d_Y$. The remote sensing instrument records this spectrum with (element-by-element) multiplicative error, represented by a diagonal error $d_{Y} \times d_{Y}$ matrix $E_{t}(\mathbf{s})$. The result is a noisy, discretized spectrum, $Y_{t}(\mathbf{s})$ ($d_{Y} \times 1$). The spectrum $Y_{t}(\mathbf{s})$ is \textit{downlinked} (i.e., transmitted from the satellite to Earth-based receiving stations) and processed through the \textit{retrieval algorithm}, $\widehat{\mathrm{R}}$. This algorithm uses a physical \textit{forward model} that captures the interaction between physical state and photons provided by the sun, or another active source. This {\em forward model} requires other inputs besides $Y_{t}(\mathbf{s})$, collectively  $\beta_{t}(\mathbf{s})$. In operations, $\beta_{t}(\mathbf{s})$ is not random, but fixed to some plausible value, $B_{t}(\mathbf{s})$. $\Gamma_{t}(\mathbf{s})$ is a random variable that represents algorithmic artifacts such as convergence to the wrong solution, etc. We leave the dimensions of $B_t(\mathbf{s})$ and $\Gamma_{t}(\mathbf{s})$ unspecified since they depend on the choice of the forward model $\hat{F}$.  

\begin{figure}[ht]
\centering
\begin{tikzpicture}
    \node[inner sep=3pt] (figA) at (0,0) {\includegraphics[width=2.75in]{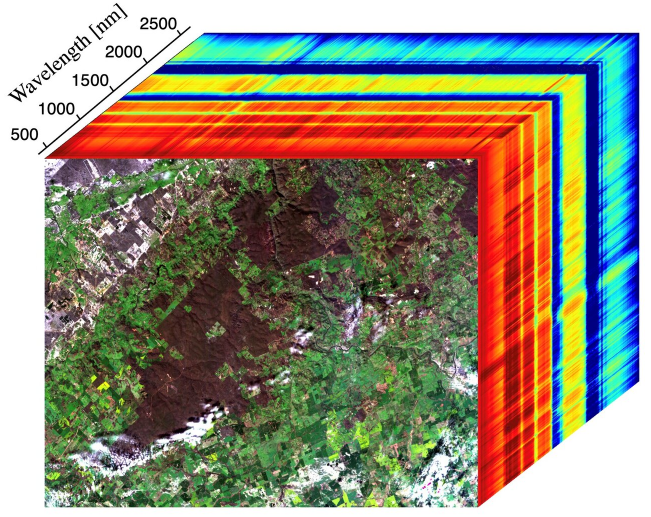}};	
    \node[right of = figA, xshift = 160, yshift = 0] (figB) {\includegraphics[width=2in]{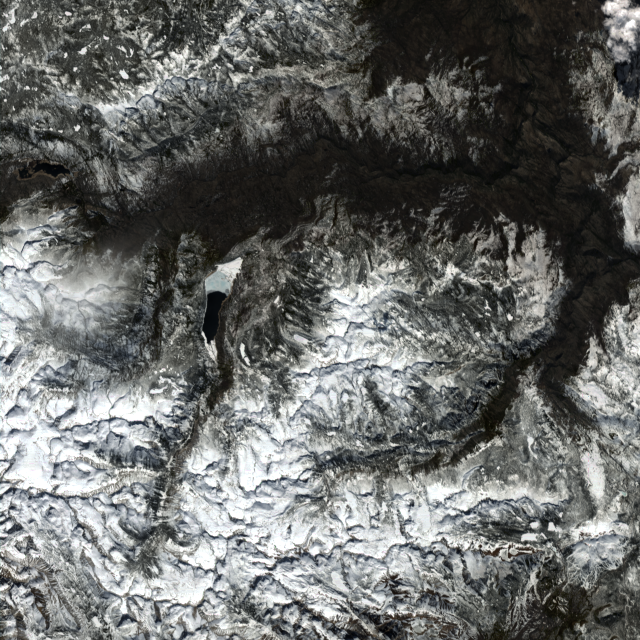}};
\end{tikzpicture}
\caption{NASA's Earth Mineral Dust Investigation (EMIT) mission. EMIT \cite{RSEThompson2024} was launched to the International Space Station in July 2022 with a primary mission of collecting data that quantifies abundance of ten surface minerals between $50^{\circ}$S and $50^{\circ}$N latitude over Africa, Asia, North and South America, and Australia. EMIT is a hyperspectral visible shortwave infrared (VSWIR) spectrometer, observing in 285 wavelength bands, and with 60-meter spatial (pixel)  resolution on the ground. EMIT data are downlinked and processed in units called ``granules", which are three-dimensional arrays of size 1280 pixels along the direction of orbit, 1240 pixels across, and 285 spectral dimensions. Left: Conceptual graphic of an EMIT granule with each of the 285 slices representing wavelength regions between 380 and 2500 nanometers. Right: RGB image of a portion (lower-left quarter) of an EMIT granule taken on March 27, 2025 over the Sierra Nevada mountains.}
\label{fig:rs4}
\end{figure}

As an example, consider NASA's Earth Mineral Dust Investigation (EMIT) mission. The satellite observes snapshots of an entire spatial field that evolves continuously over time (Figure~\ref{fig:rs4}). Here, we consider the task of uncertainty quantification for a single snapshot; therefore, we drop the subscript $t$ for the remainder of this case study. The EMIT true state vector $X(\mathbf{s})$ at a pixel $s$ is a vector of length $d_{X}=287$, comprised of a 285-dimensional {\em surface reflectance} spectrum, and two additional quantities that describe the atmosphere: water-vapor content and aerosol optical depth. 
EMIT uses a retrieval method known as optimal estimation (OE) \cite{Rodgers2000}. OE is a computational implementation of Bayes' Rule under Gaussian assumptions, and is applied to each location independently. It produces a maximum a posteriori estimate of the mode of the posterior distribution of the state given the observed spectrum, $\mathrm{P}(X(\mathbf{s})|Y(\mathbf{s}))$. The posterior mean is assumed equal to the mode, and the posterior variance is computed as a linear function of the point estimate and is widely understood to be an underestimate of the retrieval's uncertainty. See \cite{RSEThompson2018} for a thorough discussion of the details of OE. 

Sources of uncertainty affecting the quality of $\widehat{X}(\mathbf{s})$ (dimension $d_{X} \times 1$) as an estimate of $X(\mathbf{s})$ include the natural variability of $X(\mathbf{s})$, instrument measurement error $e(\mathbf{s})$, poor choices of $B(\mathbf{s})$ that can introduce bias and increase variance through interaction effects, and effects of $\Gamma(\mathbf{s})$ which are unknown. For this case study, we present a machine learning workflow that explicitly accounts for these sources of uncertainty, providing a more comprehensive estimate of uncertainty than the posterior variance of $\mathrm{P}(X(\mathbf{s})|Y(\mathbf{s}))$.

\begin{figure}[h]
    \centering
    \includegraphics[width=\linewidth]{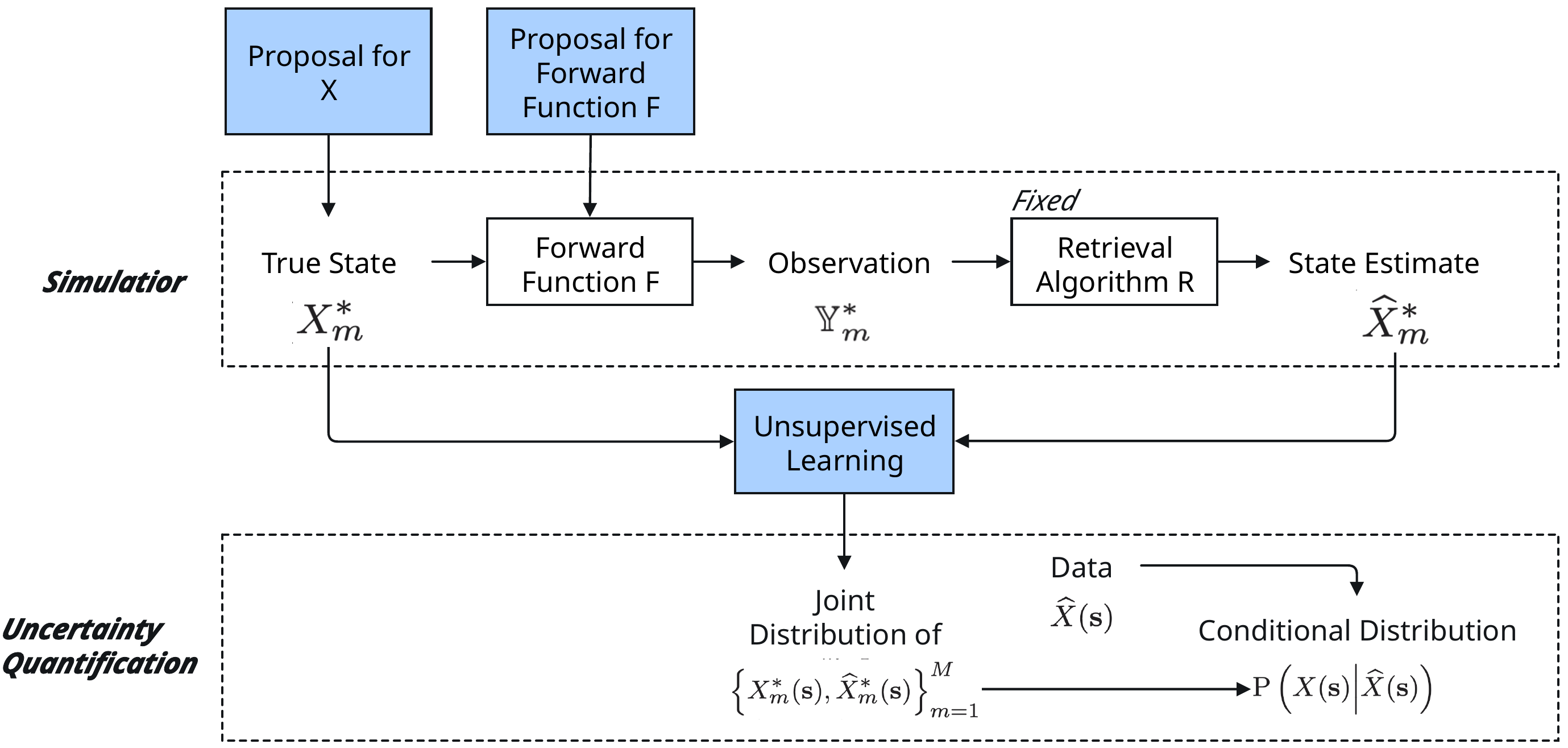}
    \caption{Simulation-based Uncertainty Quantification (UQ) with machine learning modules. This diagram shows the simulation-based framework for uncertainty quantification discussed in case study 5a, where machine learning (ML) modules are deployed at multiple key steps to enable the simulator and the inference needed for UQ. Within the simulator, machine learning is used to generate proposals for the synthetic true system state, and emulate the unobserved forward function, while a fixed retrieval algorithm reconstructs state estimates from simulated observations. An unsupervised learning module then learns a mapping between true and estimated states, enabling downstream uncertainty quantification.}
    \label{fig:remotesensing}
\end{figure}

\paragraph{Machine learning design.} In this case study, we propose an uncertainty quantification framework, shown in Figure~\ref{fig:remotesensing} for the remote sensing retrieval problem, grounded in simulation-based inference (SBI; \cite{PNASCranmerBrehmerAndLouppe2020}). The central idea is to construct a simulator that mirrors the real-world remote sensing pipeline (Figure~\ref{fig:rs1}): starting from a synthetic ``true'' state $X^*$, it generates a corresponding retrieval $\hat{X}^*$ using the same processing steps applied in practice. As shown in Figure~\ref{fig:rs1}, two key components of this pipeline are unobserved: the mechanism that generates the true state of nature, and the true forward function that maps physical states to spectra, including the effects of instrument noises. To approximate these unobserved components, we use machine learning to build a data-driven proposal model for the true state and a data-driven surrogate algorithm for the forward mapping with instrument noise (Figure~\ref{fig:remotesensing}). These machine learning modules complete the simulator that produces paired samples of true states and their corresponding retrievals. By generating a large number, $M$, of such pairs, we obtain a sample-based approximation of the joint distribution between true states and retrievals through unsupervised learning. Conditioned on a given observed retrieval, this learned joint distribution allows us to compute a well-calibrated conditional distribution over the true state, providing rigorous uncertainty quantification for the retrieved estimates.


\textbf{Machine learning module I.} The first data-driven machine learning module in Figure~\ref{fig:remotesensing} is to generate synthetic ensembles of true states from one actual instance of a retrieved spatial field. The goal is to preserve the large-scale spatial structure of the original field while introducing appropriate spatially dependent variability across realizations. Our \textit{model choice} for capturing this structure is a non-stationary Gaussian Process over the entire spatial field. This model enables the simulation of new realizations by sampling $\mathbb{X}^{\ast}_{m}$, $m=1,\ldots,M$, from it.

\begin{figure}
    \centering
    \includegraphics[width=0.7\linewidth]{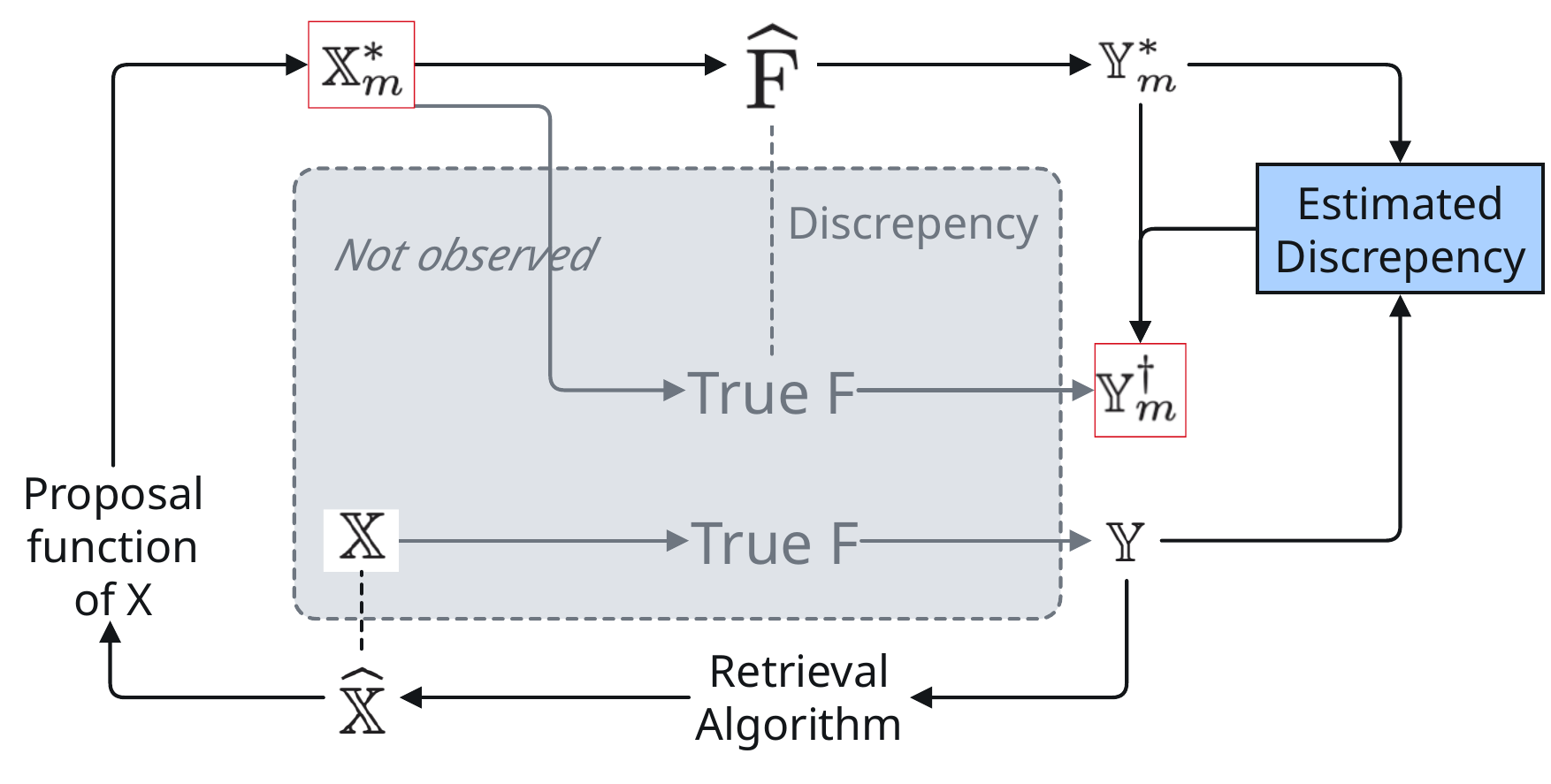}
    \caption{ML-based surrogate modeling for the forward function $\mathrm{F}$ via parametric bootstrap. Here, we illustrate the indirect strategy for simulating the unobserved forward function $\mathrm{F}$ by using a parametric bootstrap-style surrogate process. The surrogate produces $Y^*_m$ with an implicit proposal of $\mathrm{F}$ (Figure~\ref{fig:remotesensing}), emulating a comparable level of discrepancy between $F$ and $\widehat{\mathrm{F}}$.}
    \label{fig:rs-forward}
\end{figure}

\textbf{Machine learning module II.} Given each $\mathbb{X}^{\ast}_{m}$, the simulator must apply a forward function $\mathrm{F}$ to generate a corresponding observation $\mathbb{Y}^{\ast}_{m}$. As shown in Figure~\ref{fig:remotesensing}, $\mathrm{F}$ is unknown. A straightforward option is to substitute $F$ with $\widehat{\mathrm{F}}$ assumed in the retrieval algorithm. However, using $\widehat{\mathrm{F}}$ both to generate and retrieve from synthetic $\mathbb{Y}^{\ast}_{m}$ implicitly assumes that the retrieval algorithm reflects perfectly nature's true forward function. This will lead to an underestimation of uncertainty. To address this issue, we introduce a second data-driven machine learning module (Figure~\ref{fig:rs-forward}) that serves as a surrogate proposal for the true $\mathrm{F}$ by adding a sampled \textit{model discrepancy} to the synthetic spectra generated by $\widehat{\mathrm{F}}$. For each $m$, we estimate a spatial field of multiplicative {\em model discrepancies} for each $m$ as the spatial field of ratios between simulated spectra, $\mathbb{Y}^{\ast}_{m}=\hat{F}(\mathbb{X}^{\ast}_{m})$ and the true spectrum, $\mathbb{Y}$: $\mathbb{D}^{\ast}_{m}=\mathbb{Y}^{\ast}_{m} \,\, /_{<} \,\, \mathbb{Y}$. Here, the subscript ``$<$" indicates element-wise application. For each $m$, we randomly choose an index, $k_{m}$ from $\{1,\ldots,M\}$ without replacement, and we adjust $\mathbb{Y}^{\ast}_{m}$ to create $\mathbb{Y}^{\dagger}_{m}$, 
\begin{align*}
\mathbb{Y}^{\dagger}_{m} &= \mathbb{Y}^{\ast}_{m} \,\, /_{<} \,\, \mathbb{D}^{\ast}_{k_m}, \quad m=1,\ldots,M.
\label{eq:ab2}
\end{align*}
This module ensures $\left\{\mathbb{Y}^{\dagger}_{m}\right\}_{m=1}^{M}$ 
reflects the uncertainty introduced by using $\widehat{\mathrm{F}}$ in place of the true forward model.

\textbf{Machine learning module III.} Given the simulated pairs, $\left\{ X^{\ast}_{m}(\mathbf{s}), \widehat{X}^{\ast}_{m}(\mathbf{s}) \right\}_{m=1}^{M}$ at each location $\mathbf{s}$, \textit{unsupervised learning} can be applied to derive the estimated joint distribution of the synthetic true states and the retrieved state estimates, $\tilde{\mathrm{P}}\left(X^{\ast}(\mathbf{s})\Big|\widehat{X}^{\ast}(\mathbf{s}) \right)$. 
Here, we model the joint distribution of $\left\{ X^{\ast}_{m}(\mathbf{s}), \widehat{X}^{\ast}_{m}(\mathbf{s}) \right\}_{m=1}^{M}$ as a Gaussian mixture, following the procedure in \cite{JUQBraverman2021}.  

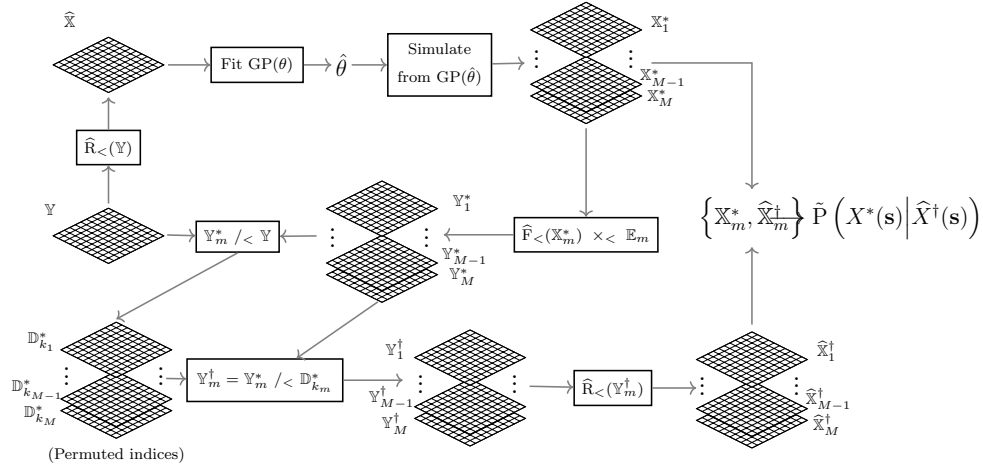
\begin{figure}[ht]
\centering
\resizebox{5.2in}{!}{%
\begin{tikzpicture}[scale=.20,every node/.style={minimum size=.35cm},on grid]
			
    \begin{scope}[local bounding box=scope1,
	xshift=-225,yshift=40, every node/.append style={
	yslant=0.5,xslant=-1},yslant=0.5,xslant=-1]
	\fill[white,fill opacity=.9] (0,0) rectangle (5,5);
	\draw[black,thin] (0,0) rectangle (5,5);
	\draw[step=5mm, black,thin] (0,0) grid (5,5);
    \end{scope}
			
    \node[above of=scope1,xshift=-20,yshift=-5] (template) {\scriptsize $\widehat{\mathbb{X}}$};									
    \node[right of=scope1,xshift=-3] (dummy1) {};
			
    \node[rectangle, draw, thick, align=center, right of=scope1,inner sep = 5, xshift=48] (Fit) 
    {\scriptsize Fit $\mathrm{GP}(\theta)$};
    
    \draw[->,thick,gray] (dummy1) -- (Fit);   	
			
    \node[align=left, right of=Fit, xshift=15, inner sep = 0] (theta) {$\hat{\theta}$};
    
    \draw[->,thick,gray] (Fit) -- (theta);   	
			
    \node[rectangle, draw, thick, align=center, right of=theta, inner sep =5, xshift = 22] (Sim) 
    {\scriptsize Simulate \\ \scriptsize from $\mathrm{GP}(\hat{\theta})$};
			
    \draw[->,thick,gray] (theta) -- (Sim);   	
			
    \begin{scope}[local bounding box=bigscope, yshift=-40]
	\begin{scope}[local bounding box=scope2, right of=Sim,
	   xshift=975,yshift=0, every node/.append style={
	   yslant=0.5,xslant=-1},yslant=0.5,xslant=-1]
	   \fill[white,fill opacity=.9] (0,0) rectangle (5,5);
       
	   \draw[black,thin] (0,0) rectangle (5,5);
	   \draw[step=5mm, black,thin] (0,0) grid (5,5);
	\end{scope}
				
	\begin{scope}[local bounding box=scope3, right of=Sim, xshift=975,yshift=30, every node/.append style={
	   yslant=0.5,xslant=-1},yslant=0.5,xslant=-1]
	   \fill[white,fill opacity=.9] (0,0) rectangle (5,5);
       
	   \draw[black,thin] (0,0) rectangle (5,5);
	   \draw[step=5mm, black,thin] (0,0) grid (5,5);
	\end{scope}
				
	\begin{scope}[local bounding box=scope4, right of=Sim,
		xshift=975,yshift=170, every node/.append style={yslant=0.5,xslant=-1},yslant=0.5,xslant=-1]
		\fill[white,fill opacity=.75] (0,0) rectangle (5,5);
        
		\draw[black,thin] (0,0) rectangle (5,5);
		\draw[step=5mm,black,thin] (0,0) grid (5,5);
	\end{scope}
    \end{scope}
			
    \node[black, left of=bigscope, xshift = 3, yshift=5] (vd1) {\large $\vdots$};
    \node[black, right of=vd1, xshift = 20, yshift=0] (vd2) {\large $\vdots$};
			
    \node[right of=bigscope, xshift = 10, yshift=20] (xstar1) {\scriptsize $\mathbb{X}^{\ast}_{1}$};
    \node[right of=bigscope, xshift = 11, yshift=-8] (xstarB) {\scriptsize $\mathbb{X}^{\ast}_{M-1}$};
    \node[right of=bigscope, xshift = 10, yshift=-18] (xstarB-1) {\scriptsize $\mathbb{X}^{\ast}_{M}$};
			
    \node[left of=bigscope,xshift=2] (dummy2) {};
    \draw[->,thick,gray] (Sim) -- (dummy2);   	
			
    \node[rectangle, draw, thick, below of=bigscope,xshift=0, yshift=-60] (F) {\scriptsize $\widehat{\mathrm{F}}_{<}(\mathbb{X}^{\ast}_{m})  \,\, \times_{<} \,\,  \mathbb{E}_{m}$};
		
    \begin{scope}[local bounding box=bigscope2,  xshift = -500, yshift=-500]
	\begin{scope}[local bounding box=scope5, right of=F, xshift=950,yshift=0, every node/.append style={
	   yslant=0.5,xslant=-1},yslant=0.5,xslant=-1]
	       \fill[white,fill opacity=.9] (0,0) rectangle (5,5);
				
            \draw[black,thin] (0,0) rectangle (5,5);
		\draw[step=5mm, black,thin] (0,0) grid (5,5);
	\end{scope}
			
	\begin{scope}[local bounding box=scope6,right of=F,xshift=950,yshift=30,every node/.append style={yslant=0.5,xslant=-1},yslant=0.5,xslant=-1]
	   \fill[white,fill opacity=.9] (0,0) rectangle (5,5);
	   \draw[black,thin] (0,0) rectangle (5,5);
	   \draw[step=5mm, black,thin] (0,0) grid (5,5);
        \end{scope}
			
	\begin{scope}[local bounding box=scope7,right of=F,xshift=950,yshift=170,every node/.append style={
    yslant=0.5,xslant=-1},yslant=0.5,xslant=-1]
	   \fill[white,fill opacity=.75] (0,0) rectangle (5,5);
		\draw[black,thin] (0,0) rectangle (5,5);
		\draw[step=5mm,black,thin] (0,0) grid (5,5);
	\end{scope}
    \end{scope}
		
    \node[black, left of=bigscope2, xshift = 3, yshift=5] (vd1) {\large $\vdots$};
    \node[black, right of=vd1, xshift = 20, yshift=0] (vd2) {\large $\vdots$};
		
    \node[right of=bigscope2, xshift = 13, yshift=20] (ystar1) {\scriptsize ${\mathbb{Y}}^{\ast}_{1}$};  
    \node[right of=bigscope2, xshift = 14, yshift=-8] (ystarB-1) {\scriptsize ${\mathbb{Y}}^{\ast}_{M-1}$};
    \node[right of=bigscope2, xshift = 13, yshift=-18] (ystarB) {\scriptsize ${\mathbb{Y}}^{\ast}_{M}$};
		
    \node[below of = bigscope2, xshift = 0, yshift =-8] (dummy8) {};
    \node[left of = F, xshift = -50, yshift = 0] (dummy9) {};
				
    \node[below of = bigscope, xshift = 0, yshift = 0] (dummy5) { };
    \draw[->, thick, gray] (dummy5) -- (F.north);
    \node[right of = F, xshift = 25, yshift = 0] (dummy6) { };
    \draw[->, thick, gray] (F) -- (dummy9);
		
    \node[rectangle, draw, thick, below of=Fit, xshift=-10, yshift=-60] (computeD) {\scriptsize $\mathbb{Y}^{\ast}_{m} \,\, /_{<} \,\, \mathbb{Y}$};	
		
		
			
	\begin{scope}[local bounding box=scope11, xshift=-225,yshift=-400, every node/.append style={	yslant=0.5,xslant=-1},yslant=0.5,xslant=-1]
	   \fill[white,fill opacity=.9] (0,0) rectangle (5,5);
	   \draw[black,thin] (0,0) rectangle (5,5);
	   \draw[step=5mm, black,thin] (0,0) grid (5,5);
	\end{scope}
			
	\node[align=center, above of=scope11,inner sep = 5, xshift=-30, yshift=-15] (Ytext) {\scriptsize $\mathbb{Y}$};
		
	\node[below of = scope11, xshift = 0, yshift = 17] (dummy21) {};	
	\node[above of = scope11, xshift = 0, yshift = -17] (dummy24) {};
	\node[below of = scope1, xshift = 0, yshift = 17] (dummy25) {};
	\node[below of = scope11, xshift = 25, yshift = 29] (dummy31) {};
	\node[left of = bigscope2, xshift = 0, yshift = 3] (dummy32) {};
		
	\draw[->, thick, gray] (dummy32) -- (computeD);
	\draw[->, thick, gray] (dummy31) -- (computeD);	
		
	\node[above of = scope11, rectangle, draw, thick, xshift = 0, yshift = 17] (R0) {\scriptsize $\widehat{\mathrm{R}}_{<}(\mathbb{Y})$};

	\draw[->, thick, gray] (dummy24) -- (R0);		
	\draw[->, thick, gray] (R0) -- (dummy25);			
	\begin{scope}[local bounding box=bigscope4,  xshift = -585, yshift=-850]
	   \begin{scope}[local bounding box=scope21,xshift=380,yshift=0,every node/.append style={yslant=0.5,xslant=-1},yslant=0.5,xslant=-1]
		  \fill[white,fill opacity=.9] (0,0) rectangle (5,5);
		  \draw[black,thin] (0,0) rectangle (5,5);
		\draw[step=5mm, black,thin] (0,0) grid (5,5);
	\end{scope}
			
	\begin{scope}[local bounding box=scope22,xshift=380,yshift=30,every node/.append style={yslant=0.5,xslant=-1},yslant=0.5,xslant=-1]
		\fill[white,fill opacity=.9] (0,0) rectangle (5,5);
		\draw[black,thin] (0,0) rectangle (5,5);
		\draw[step=5mm, black,thin] (0,0) grid (5,5);
	\end{scope}
			
	\begin{scope}[local bounding box=scope23,below of=F,xshift=380,yshift=185,every node/.append style={yslant=0.5,xslant=-1},yslant=0.5,xslant=-1]
	   \fill[white,fill opacity=.75] (0,0) rectangle (5,5);
	   \draw[black,thin] (0,0) rectangle (5,5);
	   \draw[step=5mm,black,thin] (0,0) grid (5,5);
	\end{scope}
    \end{scope}
		
    \node[black, left of=bigscope4, xshift = 3, yshift=5] (vd3) {\large $\vdots$};
    \node[black, right of=vd3, xshift = 20, yshift=0] (vd4) {\large $\vdots$};
		
    \node[left of=bigscope4, xshift = -10, yshift=20] (dstar1) {\scriptsize $\mathbb{D}^{\ast}_{k_{1}}$};
    \node[left of=bigscope4, xshift = -12, yshift=-5] (dstarB-1) {\scriptsize $\mathbb{D}^{\ast}_{k_{M-1}}$};
    \node[left of=bigscope4, xshift = -10, yshift=-18] (dstarB) {\scriptsize $\mathbb{D}^{\ast}_{k_{M}}$};

    \node[right of=bigscope4, draw, thick, rectangle, xshift = 48, yshift = 0, inner sep = 5pt] (Ydagger) {\scriptsize $\mathbb{Y}^{\dagger}_{m} = \mathbb{Y}^{\ast}_{m} \,\, /_{<} \,\, \mathbb{D}^{\ast}_{k_{m}}$};
		
    \node[below of = bigscope2, xshift = 3, yshift = 1] (dummy91) { };
    \draw[->, thick, gray] (dummy91) -- (Ydagger);
		
    \node[below of=bigscope4, xshift = 0, yshift = -10] (perm) {\scriptsize (Permuted indices)}; 
		
    \node[above of=bigscope4, xshift = 0, yshift = 3] (dummy41) {};
    \node[right of=bigscope4, xshift = -8, yshift = 1] (dummy42) {};
		
    \draw[->, thick, gray] (computeD.south) -- (dummy41.center);
    \draw[->, thick, gray] (dummy42) -- (Ydagger);
		
    \begin{scope}[local bounding box=bigscope5,  xshift = 325, yshift=-870]
	\begin{scope}[local bounding box=scope31,xshift=380,yshift=0,every node/.append style={yslant=0.5,xslant=-1},yslant=0.5,xslant=-1]
	   \fill[white,fill opacity=.9] (0,0) rectangle (5,5);
	   \draw[black,thin] (0,0) rectangle (5,5);
	   \draw[step=5mm, black,thin] (0,0) grid (5,5);
	\end{scope}
			
	\begin{scope}[local bounding box=scope32,xshift=380,yshift=30,every node/.append style={yslant=0.5,xslant=-1},yslant=0.5,xslant=-1]
	   \fill[white,fill opacity=.9] (0,0) rectangle (5,5);
	   \draw[black,thin] (0,0) rectangle (5,5);
	   \draw[step=5mm, black,thin] (0,0) grid (5,5);
        \end{scope}
			
	\begin{scope}[local bounding box=scope33,below of=F,xshift=380,yshift=185,every node/.append style={yslant=0.5,xslant=-1},yslant=0.5,xslant=-1]
            \fill[white,fill opacity=.75] (0,0) rectangle (5,5);
		  \draw[black,thin] (0,0) rectangle (5,5);
		  \draw[step=5mm,black,thin] (0,0) grid (5,5);
	\end{scope}
    \end{scope}
		
    \node[black, left of=bigscope5, xshift = 3, yshift=5] (vd31) {\large $\vdots$};
    \node[black, right of=vd31, xshift = 20, yshift=0] (vd41) {\large $\vdots$};
		
    \node[left of=bigscope5, xshift = -10, yshift=20] (ydagger1) {\scriptsize $\mathbb{Y}^{\dagger}_{1}$};
    \node[left of=bigscope5, xshift = -12, yshift=-5] (ydaggerB-1) {\scriptsize $\mathbb{Y}^{\dagger}_{M-1}$};
    \node[left of=bigscope5, xshift = -10, yshift=-18] (ydaggerB) {\scriptsize $\mathbb{Y}^{\dagger}_{M}$};

    \node[left of=bigscope5, xshift = -1, yshift = 4] (dummy51) {};
    \node[right of=bigscope5, xshift = -3, yshift = 1] (dummy52) {};
		
    \draw[->, thick, gray] (Ydagger) -- (dummy51);
			
    \node[right of = bigscope5, rectangle, draw, thick, xshift = 45, yshift = 0] (R) {\scriptsize $\widehat{\mathrm{R}}_{<}(\mathbb{Y}^{\dagger}_{m})$};
		
    \draw[->, thick, gray] (dummy52) -- (R);
		
    \begin{scope}[local bounding box=bigscope6,  xshift = 1050, yshift=-875]
    
	\begin{scope}[local bounding box=scope61,xshift=380,yshift=0,every node/.append style={yslant=0.5,xslant=-1},yslant=0.5,xslant=-1]
	   \fill[white,fill opacity=.9] (0,0) rectangle (5,5);
	   \draw[black,thin] (0,0) rectangle (5,5);
	   \draw[step=5mm, black,thin] (0,0) grid (5,5);
        \end{scope}
			
	\begin{scope}[local bounding box=scope62,xshift=380,yshift=30,every node/.append style={yslant=0.5,xslant=-1},yslant=0.5,xslant=-1]
	   \fill[white,fill opacity=.9] (0,0) rectangle (5,5);
	   \draw[black,thin] (0,0) rectangle (5,5);
	   \draw[step=5mm, black,thin] (0,0) grid (5,5);
	\end{scope}
			
	\begin{scope}[local bounding box=scope63,below of=F,xshift=380,yshift=185,every node/.append style={yslant=0.5,xslant=-1},yslant=0.5,xslant=-1]
	   \fill[white,fill opacity=.75] (0,0) rectangle (5,5);
	   \draw[black,thin] (0,0) rectangle (5,5);
	   \draw[step=5mm,black,thin] (0,0) grid (5,5);
	\end{scope}
    \end{scope}
		
    \node[black, left of=bigscope6, xshift = 3, yshift=5] (vd5) {\large $\vdots$};
    \node[black, right of=vd5, xshift = 20, yshift=0] (vd6) {\large $\vdots$};
		
    \node[right of=bigscope6, xshift = 10, yshift=20] (hatxstar1) {\scriptsize $\widehat{\mathbb{X}}^{\dagger}_{1}$};
    \node[right of=bigscope6, xshift = 11, yshift=-5] (hatxstarB-1) {\scriptsize $\widehat{\mathbb{X}}^{\dagger}_{M-1}$};
    \node[right of=bigscope6, xshift = 10, yshift=-18] (hatxstarB-1) {\scriptsize $\widehat{\mathbb{X}}^{\dagger}_{M}$};
		
    \node[above of=bigscope6, xshift = 0, yshift = 0] (dummy61) {};
    \node[left of=bigscope6, xshift = 5, yshift = 1] (dummy62) {};
		
    \draw[->, thick, gray] (R) -- (dummy62);
		
    \node[above of = bigscope6, xshift = 0, yshift = 60] (ens) {$\left\{ \mathbb{X}^{\ast}_{m}, \widehat{\mathbb{X}}^{\dagger}_{m}\right\}$};
		
    \draw[->, thick, gray] (dummy61) -- (ens);
		
    \node[below of=ens, xshift = 0, yshift = 20] (dummy71) {};
    \node[above of=ens, xshift = 0, yshift = -20] (dummy72) {};
    \node[right of=bigscope, xshift = 0, yshift = 0] (dummy80) {};
		
    \draw[->,thick, gray] (dummy80) -| (ens);

    \node[right of=ens, xshift=35, yshift=0] (cprob) {$\longrightarrow \tilde{\mathrm{P}}\left( X^{\ast}(\mathbf{s})\Big| \widehat{X}^{\dagger}(\mathbf{s})  \right)$};

\end{tikzpicture}
} 
\caption{Conceptual graphic of the UQ workflow for an EMIT scene. Quantities in blackboard font indicate full spatial fields. Any operation subscripted by ``$<$" indicates element-wise application.}
\label{fig:rs5}
\end{figure}

\paragraph{Machine learning development and deployment.} 
Combining all the design choices, Figure~\ref{fig:rs5} presents the graphical diagram of the entire uncertainty quantification workflow for this case study. The flow begins on the left with a spatial field of actual spectra, $\mathbb{Y}$, which is a ``data cube" (a three-dimensional array) of EMIT data similar to the left side of Figure~\ref{fig:rs4}. 
Every pixel in $\mathbb{Y}$ is processed through the operational retrieval algorithm, $\widehat{\mathrm{R}}$ with all other arguments and settings assigned fixed values, but shown here without those to save space. The output is the operationally retrieved estimated spatial field of surface reflectances, $\widehat{\mathbb{X}}$.  Together, $\mathbb{Y}$ and $\widehat{\mathbb{X}}$ serve as data to drive the proposed simulation-based framework. 

To fit a non-stationary Gaussian process model (machine learning module I) to the field $\widehat{\mathbb{X}}$, we first preprocess the data by projecting each 285-dimensional vector $\widehat{X}(\mathbf{s})$ into the space of the leading six components (captures almost 100 percent of the variation) of the covariance matrix, $\mathrm{cov}(\widehat{\mathbf{X}})$. This step leads to more tractable computation and stable modeling results. This reduces the input field from a $640 \times 640 \times 285$ array to a more manageable $640 \times 640 \times 6$ array. For uncertainty quantification, the simulation strategy is guided by which sources of variability we aim to capture. In the example of one single snapshot for EMIT, we treat the first two principal components, which are associated with large-scale trends, as fixed, and model the variability in the remaining components using non-stationary Gaussian processes. 
We simulate $M$ realizations of each independently, and reverse the principal component transformation to obtain realizations in the original 285-dimensional space. 
Simulated realizations are denoted by $\mathbb{X}^{\ast}_{m}$ in Figure~\ref{fig:rs5}. 

One computational intensive step of this workflow concern the retrieval on every pixel in every member of the ensemble $\left\{\mathbb{Y}^{\dagger}_{m}\right\}_{m=1}^{M}$ in exactly the same way as it is done in operations. This can be a slow process since it is $M$-times as expensive as the actual retrieval. Ongoing work explores replacing it with a retrieval emulator that replicates the retrieval’s behavior with known uncertainty characteristics \cite{JUQSusiluotoEtAl2025+}. 

\subsection*{Case Study 2b: Probabilistic Programming for Data-Driven Parameter Inference}
\paragraph{Scientific problem.} In climate modeling, \textit{data assimilation} (DA) is traditionally used to adjust model state trajectories to better match observations, improving short-term forecasts and physical realism. However, this state-centric approach is less effective for long-term climate projections, where the goal is to capture the distribution of states (such as the frequency and intensity of extreme events) rather than specific trajectories. At the same time, model tuning, i.e., calibrating uncertain physical parameters using observations, is often a manual, heuristic process. Existing DA and tuning methods typically treat state correction and model improvement as separate tasks. This is partly due to the fact that tuning based on matching long unrolled simulations with observations is computationally expensive and error-prone \cite{hannart2016,carrassi2017a,metref2019}. Data assimilation that improves state estimates without addressing structural model biases represents a missed opportunity to use observational data not just to correct model output, but to inform and improve the model itself. Machine learning, particularly generative models, offers a new opportunity to treat DA and model tuning as a joint inverse problem. Instead of only estimating system states, we can seek to infer the physical model parameters that most likely explain the observed data distributions. To address this challenge, Li et al.\ (2025) \cite{li2025probabilistic} offers a proof-of-concept of such as setup using the Lorenz '96 model. The Lorenz '96 model's inherent chaotic dynamics provide a suitable test case for predictability and data assimilation studies.

\begin{figure}[h]
\centerline{\includegraphics[angle=0,width=\linewidth]{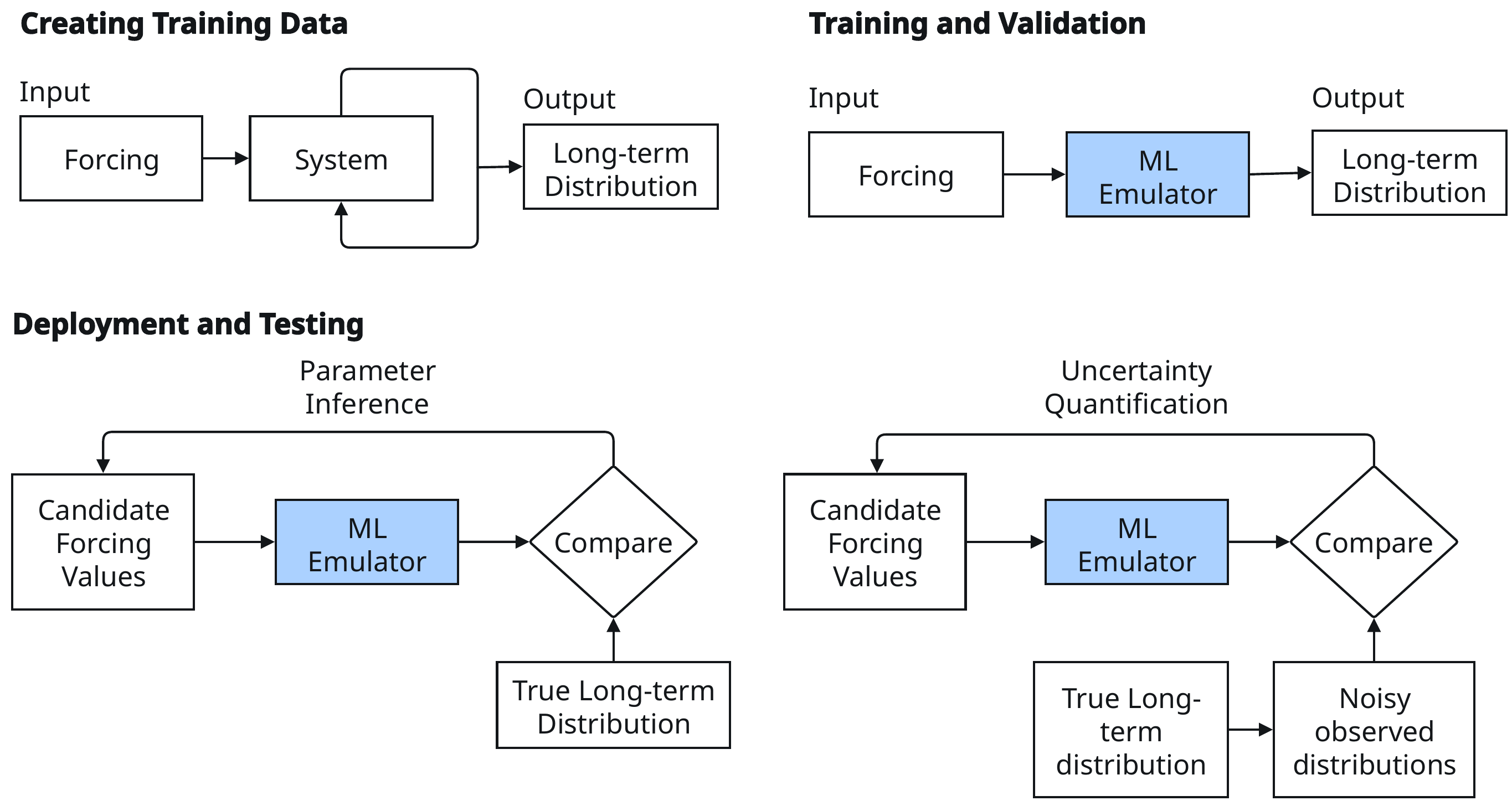}}
\caption{Workflow for distribution-based parameter inference using probabilistic machine learning. In the training phase (top), input–output pairs are generated from a physical model simulator under various forcing parameters and used to train a machine learning model that emulates the long-term distribution of climate states. In the deployment phase (bottom), the trained emulator is used in reverse: given an observed distribution and a prior over physical parameters, the model infers a posterior distribution over forcings via probabilistic programming.}
\label{fig:cnf}
\end{figure}

\paragraph{Machine learning design.}  Li et al.\ (2025) \cite{li2025probabilistic} propose a probabilistic programming framework ~\cite{PNASCranmerBrehmerAndLouppe2020} that re-imagines data assimilation as a form of data (distribution)-driven parameter inference (model tuning). The core idea is to first learn a reversible generative model (e.g., a conditional normalizing flow \cite{winkler2019learning}) that maps model parameters to observed data: the probability density of observable state variables. Instead of learning to reproduce specific time series trajectories, the machine learning model is trained to emulate the distributional behavior of the system. Once trained, this machine learning distribution emulator can then enable the inverse inference: given observed distributions (e.g., temperature frequency over time), one can infer the most plausible physical parameter values (Figure~\ref{fig:cnf}) with uncertainty, taking advantage of the fast generation process of generative machine learning models in contrast to the conventional ensemble-based climate models.

In \cite{li2025probabilistic}, the \textit{training data} are simulated using, a simplified system, the Lorenz '96 (L96) system with 4 large-scale variables and 32 small-scale variables. 500 pairs of varying forcing ($F \in [10,50]$; input), and corresponding long-term distributions (output) are generated. 

The \textit{model choice} for a machine learning distribution emulator based on training data in \cite{li2025probabilistic} was \textit{conditional normalizing flow} models, which enables the replication of state distributions without explicitly generating the complete trajectory series. 
Normalizing flows (NFs) are generative models that transform simple base distributions (e.g., Gaussian) into complex data distributions through invertible transformations. The probability density function under transformation $x = g(z)$ is:
\begin{equation*}
p_X(x) = p_Z(g^{-1}(x)) \left| \frac{d}{dx}g^{-1}(x) \right|,
\end{equation*}
where $p_Z(z)$ is the base distribution.

\textit{Conditional normalizing flows} extend NFs by conditioning transformations on external parameters $\theta$ (e.g., $F$). Thus, the conditional probability density function is given by:
\begin{equation*}
p_X(x, \theta) = p_Z(z, \theta) \left| \frac{\partial z}{\partial x} \right|,
\end{equation*}
where $z=f(x, \theta)$ integrates conditional parameters into the transformation.

For the \textit{objective function}, the cNF parameters are trained by minimizing the negative log-likelihood of observed state distributions:
\begin{equation*}
\mathcal{L} = -\sum_{i=1}^{n} \log p_X(x_i) = -\sum_{i=1}^{n}\left[\log p_Z(z_i, \theta_i) + \mathcal{NN}(x_i, \theta_i)\right],
\end{equation*}
where $\mathcal{NN}(x, \theta)$ is the neural network parameterizing the transformations.

\paragraph{Model development.}
Model development in \cite{li2025probabilistic} follows standard supervised learning procedures, made straightforward by the modular structure of the dataset and the well-defined machine learning design. Once the architecture and conditioning variables were selected, the model was trained using established deep learning workflows with minimal task-specific modification.

\paragraph{Model deployment and evaluation.} The trained machine learning distribution emulator is deployed within a probabilistic inference framework to estimate model parameters from observed data (Figure~\ref{fig:cnf}). Rather than simulating the full trajectories of the system, the emulator predicts the probability density of the state variables conditioned on the forcing parameters $F$, allowing efficient inference of parameter parameters based on the distribution.

\textit{Parameter inference} is cast as an optimization problem: we identify $F_*$ that minimizes the discrepancy between the emulator output $\hat{p}_X(\hat{X}|F)$ and the observed distribution $p_X(x)$. This discrepancy is measured using Maximum Mean Discrepancy (MMD) \cite{fortet1953convergence, borgwardt2006integrating}, a kernel-based metric that quantifies distributional similarity directly from sample data. We compute MMD using an RBF kernel and identify $F$ through grid search:
\begin{equation}
F_* = \arg\min_F \left\{ -\widehat{\text{MMD}}\left[ \hat{p}_X(\hat{X}|F), p_X(X) \right] \right\}.
\end{equation}

\textit{Uncertainty quantification (UQ)} is performed by using multiple noisy {\em observed} empirical distributions, $\left\{p_X^{(i)}(X), i=1, ..., B\right\}$ drawn from the true long-term distribution and repeating the inference across these samples. This procedure results in a distribution of inferred $F_*$ values:
\begin{equation}
F_*^{(i)} = \arg \min_F \left\{ -\widehat{\text{MMD}} \left[ \hat{p}_X^{(i)}(\hat{X}_k|F),  p_X^{(i)}(X_k) \right] \right\},
\end{equation}
providing a confidence set for the inferred parameter.

Their framework has been validated on three test cases with known $F$ values. The deterministic approach successfully recovers $F_*$ with minimal error, while the probabilistic approach yields tight uncertainty bounds around the true values. Li et al.\ (2025) \cite{li2025probabilistic} confirmed that the inferred distributions closely match the observations, demonstrating the effectiveness of distribution-based parameter inference.


The probabilistic data assimilation framework developed in \cite{li2025probabilistic} can further be applied to scientific questions of interest, such as estimating the likelihood of extreme events and characterizing joint behavior of distributional extremes. The trained machine learning distribution emulator is used to estimate the probability of specific outcomes, such as extreme values falling within a given range, while accounting for uncertainty in the underlying physical forcing. In addition, the framework supports joint quantile analysis to assess the likelihood of compound extremes (e.g., simultaneous high and low anomalies), which is critical for risk assessment \cite{li2025probabilistic}.

\section{ML-First Case Studies}
\label{sec:mlfirst}

\subsection*{Case Study 3a: Benchmarking ML Models on Subseasonal-to-Seasonal Forecasting Tasks with ChaosBench}

\paragraph{Scientific problem.} Subseasonal-to-seasonal (S2S) forecasting is one of the most challenging problems in climate science. It sits between short-term weather prediction and long-term climate projections. At this timescale, prediction accuracy depends strongly on both initial conditions and boundary conditions \cite{PriveErrico2013,WuLynchRivers2005,Lorenz1963}
, as well as complex coupled interactions across the atmosphere, ocean, and land surface. Traditionally, physics-based models are used to generate forecasts by numerically solving the governing equations of the Earth system. They produce globally consistent forecasts with relatively limited observational data, but are computationally expensive and can be difficult to tune or adapt to new regimes. 

\begin{figure}[h]
    \centering
    \includegraphics[width=0.65\linewidth]{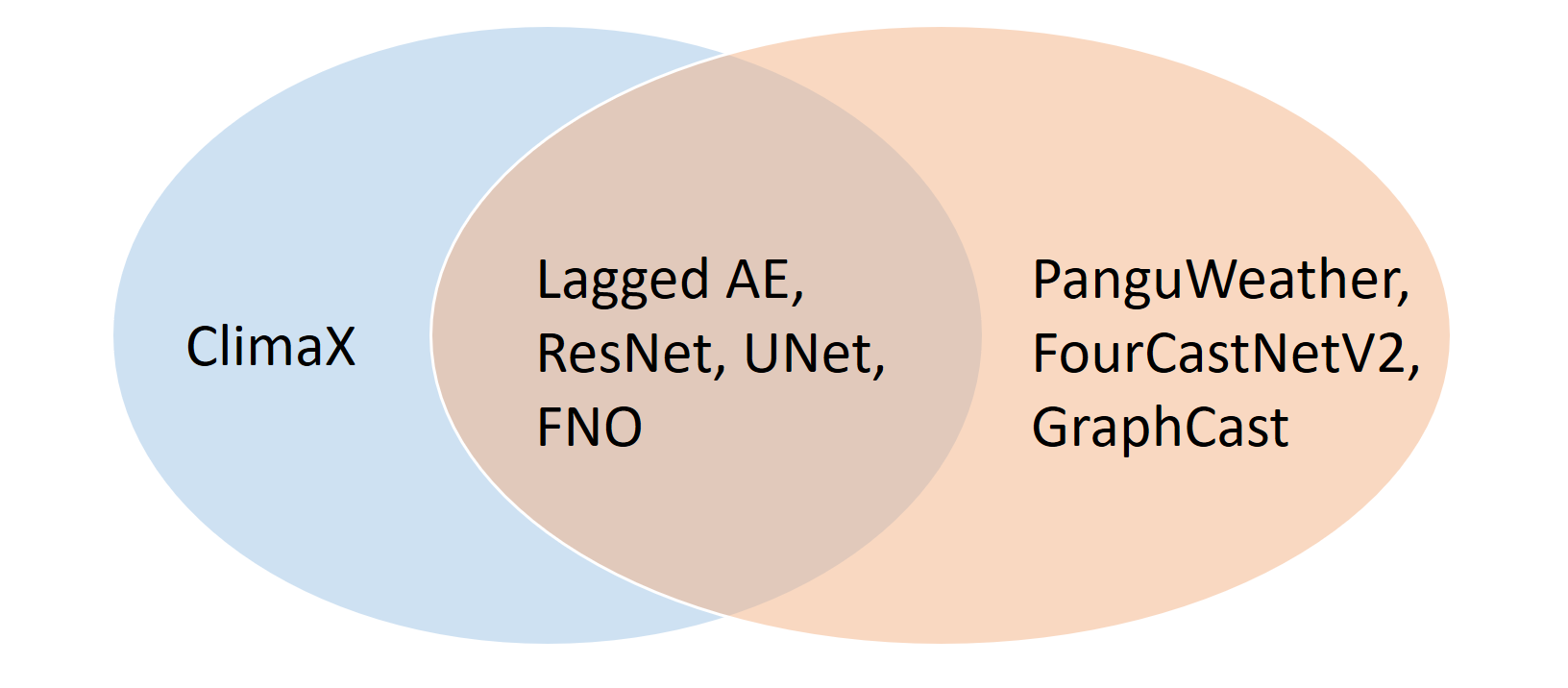}
    \caption{Model groups evaluated in ChaosBench project \cite{nathaniel2024chaosbench}. Blue: models used only with direct forecasting (e.g., ClimaX). Orange: models used in autoregressive mode only (e.g., PanguWeather, GraphCast). Overlapping region: general-purpose models applied under both strategies.}
    \label{fig:model-venn}
\end{figure}

\paragraph{Machine learning design.} Nathaniel et al. (2024) \cite{nathaniel2024chaosbench} carried out the \textit{ChaosBench} project to evaluate the predictive skills of machine learning emulators in the sub-seasonal to seasonal range. Let $\mathbf{X}_t \in \mathbb{R}^{H \times W \times C}$ denote the full state of the system  at day $t$, where $H$, $W$, and $C $ represent the spatial height, width, and number of climate variables, respectively. The goal is to predict future states $\mathbf{X}_{t+\tau}$ from a given input state $\mathbf{X}_t $, where $\tau$ denotes the lead time, i.e., the number of days ahead to forecast. 
%
%
Machine learning {\em model choices} can be considered in two ways:
\begin{itemize}
    \item \textit{Direct model:}, a model $G_{\phi_\tau}$ is trained for each target lead time $\tau \in \{1, 5, 10, \dots, 44\}$: 
    \[
    \hat{\mathbf{X}}_{t+\tau} = G_{\phi_\tau}(\mathbf{X}_t).
    \] 
This avoids the propagation of recursive errors but comes at the cost of training multiple models.
  \item \textit{Autoregressive model:}, a model $F_\theta$ is trained to perform one-step-ahead predictions and be applied repeatedly:
      \[
      \hat{\mathbf{X}}_{t+1} = F_\theta(\mathbf{X}_t), \quad \hat{\mathbf{X}}_{t+2} = F_\theta(\hat{\mathbf{X}}_{t+1}), \dots
      \hat{\mathbf{X}}_{t+\tau} = F_\theta(\hat{\mathbf{X}}_{t+\tau-1}).
      \]
This strategy mimics the way physics-based dynamical models evolve, but suffers from error accumulation over long horizons. 
\end{itemize}
%
While most predictive machine learning models can, in principle, be applied in both settings, practical considerations often make a model more suitable for one than the other. The \textit{autoregressive strategy} is training-efficient and naturally aligned with physics-based simulation paradigms, but it incurs higher forecasting costs due to its need for sequential roll-out. In contrast, the \textit{direct strategy} trains separate models for each forecast horizon $\tau$, leading to increasing computational costs for data preparation as $\tau$ grows. We visualize in Figure~\ref{fig:model-venn} models evaluated in the Chaosbench project \cite{nathaniel2024chaosbench}.
\begin{itemize}
    \item \textit{Autoregressive only}:  Pangu-Weather \cite{PanguBiEtAl2022}, FourCastNetV2 \cite{pathak2022fourcastnet}, and GraphCast \cite{GraphCastLamEtAl2022} are pre-trained numerical weather prediction surrogates designed to emulate the stepwise evolution of the atmosphere. They are readily applicable under the autoregressive setting. Adapting these models for the direct long-range forecasting setting would require substantial architectural redesign and dense, multi-horizon supervision, leading to significantly higher training costs in both data volume and compute. 
    \item \textit{Direct only}: ClimaX \cite{ClimaXNguyenEtAl2023} is a transformer-based foundation model pre-trained in CMIP6 simulations, where the lead time is explicitly encoded in the input. This architectural design enables flexible horizon prediction without recursive rollout, making it naturally suited for the direct strategy.
    \item \textit{Both settings}: Lagged Autoencoder \cite{LaggedAEChenEtAl2022}, ResNet \cite{ResNetRaspThuerey2021}, UNet \cite{UnetRaspEtAl2020}, and FNO \cite{FNOLiEtAl2020} are general-purpose architectures with relatively light-weight training requirements, allowing easy adaptation to both strategies.
\end{itemize}
%
%
The {\em training data} of the chaosbench project were sourced from the ERA5 reanalysis data \cite{ERA5HersbachEtAl2020}, covering the period from 1979 to 2023. 
%
The construction of training input-output pairs depends on the forecasting strategy (autoregressive or direct). All models use mean squared error (MSE) as the \textit{loss function} during training. 
%

\paragraph{Model development.} During training, all models were trained on data from 1979-2015, validated in 2016-2021, and tested in 2022-2023. Most models are randomly initialized and trained from scratch. The exceptions were (1) \textit{ClimaX}, which was initialized from a ViT model pretrained on CMIP6 simulations \cite{CMIP6EyringEtAl2016}; and (2) the large-scale autoregressive only models, (PanguWeather, GraphCast, FourCastNetV2), were applied directly using their released pretrained checkpoints for inference without additional training.

\paragraph{Model deployment and evaluation.} All models were deployed to make forecasts on the test data. In addition to MSE, the forecasts were evaluated using vision-based metrics to assess accuracy and similarity with ground truth, including bias, anomaly correlation coefficient (ACC) \cite{ACCMurphyEpstein1989}, and MS-SSIM \cite{MS-SSIMWangEtAl2003}. The ChaosBench project further introduced two metrics in the spectral domain that focus on high frequency information: \textit{ spectral divergence (SpecDiv)} adapted from KL divergence and measures the difference in power spectra between prediction and ground truth; and \textit{spectral residual (SpecRes)} that computes the RMSE between normalized spectral densities.

\subsection*{Case Study 3b: Physics-Informed Transfer Learning for Spatiotemporal Upscaling of Sparse Air-Sea pCO2 Observations}

\paragraph{Scientific Problem.}  
Sparse climate observations limit our ability to monitor key processes and benchmark Earth system models \cite{righi2020earth}. A representative challenge is upscaling partial pressure of carbon dioxide ($pCO_2$), where observations cover only 1--2\% of the ocean \cite{sabine2013surface,bakker2014update}. Traditional data assimilation approaches \cite{strebel2022coupling,qu2024deep} are hindered by model structural errors \cite{anav2015spatiotemporal,nathaniel2024chaosbench}, while direct machine learning methods \cite{jung2019fluxcom,chen2019machine} extrapolate poorly into undersampled regions like the Southern Ocean, producing biased and inconsistent estimates \cite{gruber2023trends}.

\paragraph{Machine Learning Design.} Kim et al.\ (2024) \cite{kim_spatiotemporal_2024} formulates a solution as as a supervised regression problem that predicts (imputes) monthly oceanic partial pressure of CO$_2$ ($pCO_2$) at $1^\circ \times 1^\circ$ resolution. Each prediction is based on a set of geophysical inputs: sea surface temperature (SST) \cite{reynolds2002improved}, surface chlorophyll-a (CHL) \cite{maritorena2010merged}, sea surface salinity (SSS) \cite{good2013en4}, atmospheric CO$_2$ mixing ratio (xCO$_2$) \cite{conway1994evidence}, and distance to coast (D2C). Latitude and longitude are also included as features. The resulting tabular dataset treats each ocean pixel and time point as an individual training sample with seven inputs and one $pCO_2$ target. 

\textit{Model choice}: to capture the spatial and spatiotemporal structures in $pCO_2$ fields, Kim et al.\ (2024) \cite{kim_spatiotemporal_2024} combines two convolutional deep learning models: U-NET \cite{ronneberger2015u} and ConvLSTM \cite{shi2015convolutional}. U-NET employs an encoder-decoder architecture to learn multiscale spatial features, while ConvLSTM integrates convolutional operations within recurrent units to capture temporal dependencies across monthly inputs. 


\textit{Training data} were monthly $pCO_2$ data from 1982 to 2017 using two sources: 1) simulated outputs from four global ocean biogeochemistry models (GOBMs) in the Large Ensemble Testbed \cite{gloege2022improved,gloege2021quantifying} and sparse in-situ observations from the SOCAT dataset \cite{bakker2014update}.

\paragraph{Model development.} Kim et al.\ (2024) \cite{kim_spatiotemporal_2024} adopted a two-phase training strategy based on transfer learning. First, models are pre-trained on dense simulated $pCO_2$ data from GOBM ensemble members (e.g., CESM, MPI, CanESM), using five members for training and one for validation. U-NET is trained to reconstruct $pCO_2$ fields from input features, and its outputs are appended as an extra channel to the ConvLSTM inputs. ConvLSTM receives input sequences of $K$ months and predicts the next $K$ steps, capturing temporal dynamics. Both models are trained using the Adam optimizer \cite{kingma2014adam} with a learning rate of $10^{-3}$, ELU activations \cite{clevert2015fast}, and batch size of 16. Hyperparameters are selected via randomized search with 7-fold cross-validation \cite{bergstra2012random}. The final ConvLSTM has about 1.1M parameters.
%


In the fine-tuning phase, models are updated using sparse SOCAT $pCO_2$ observations, with losses computed only at observed locations. The same two-step structure is used: fine-tune U-NET, then ConvLSTM. To retain generalization and avoid overfitting, only 2--5\% of parameters are unfrozen, and early stopping is applied. This allows the model to integrate sparse observations while preserving physically informed spatiotemporal structure.

\paragraph{Model evaluation and validation.}  
Kim et al.\ (2024) \cite{kim_spatiotemporal_2024} evaluated model performance using RMSE, with MSE as the loss function during training. 
%
Model performance was assessed on two fronts: \textit{interpolation} (held-out SOCAT tracks) and \textit{extrapolation} (unseen GOBM members). These evaluations test how well models generalize both within and beyond sparse observations.

\bibliographystyle{plainnat}
\bibliography{references} 

\end{document}